\pdfoutput=1
\documentclass{article}
\usepackage{iclr2027_conference}
\usepackage[T1]{fontenc}
\renewcommand{\sfdefault}{phv}
\usepackage[T1]{fontenc}

\usepackage{amsmath,amsfonts,bm}

\def\eqref#1{equation~\ref{#1}}

\def\1{\bm{1}}

\DeclareMathAlphabet{\mathsfit}{\encodingdefault}{\sfdefault}{m}{sl}
\SetMathAlphabet{\mathsfit}{bold}{\encodingdefault}{\sfdefault}{bx}{n}

\usepackage{hyperref}
\hypersetup{hidelinks}
\usepackage{url}
\usepackage{graphicx}
\usepackage{booktabs}
\usepackage{array,tabularx}
\usepackage{microtype}
\usepackage{seqsplit}
\usepackage{tikz}
\usetikzlibrary{arrows.meta,positioning,fit}

\title{Audit-First VAPO: Risk-Certified Selective\\Updates under Imperfect Verification}
\author{Miaobo Hu$^{1,2,*}$, Shuhao Hu$^{2}$, Xiaobo Guo$^{2}$, Xin Wang$^{2}$,\\
Bokun Wang$^{2}$, Rui Chen$^{2}$, Daren Zha$^{2}$\\[4pt]
{\normalfont $^{1}$School of Artificial Intelligence, University of Chinese Academy of Sciences, Beijing, China}\\
{\normalfont $^{2}$Institute of Information Engineering, Chinese Academy of Sciences, Beijing, China}\\
{\normalfont $^{*}$Corresponding author: \texttt{humiaobo@iie.ac.cn}}}
\newcolumntype{P}[1]{>{\raggedright\arraybackslash}p{#1}}

\iclrfinalcopy
\hypersetup{pdfauthor={Miaobo Hu, Shuhao Hu, Xiaobo Guo, Xin Wang, Bokun Wang, Rui Chen, Daren Zha}}
\begin{document}
\maketitle
\lhead{Preprint}

\begin{abstract}
Imperfect verifiers can assign a harmful update direction even when clipping
and regularization bound its magnitude. We introduce \emph{Audit-First VAPO},
which separates discrete directional admission from continuous magnitude
control. An observation-only accept--appeal--abstain policy uses a finite
secondary-verification budget; its action trace is frozen before clean labels
are joined. Simultaneous finite-sample bounds then certify selected harmful
risk, coverage, and verifier-call rate over a predeclared policy family.
Conditional Hoeffding--Azuma bounds account for the dependence induced by
shared budgets, and rollout or verifier changes initiate a new certification
stage. After admission, a bounded trust--clip--KL actuator controls magnitude.
We evaluate two models on two reasoning benchmarks against static RLVR,
matched-random selection, confidence thresholding, noise correction, and
verifier augmentation. On Qwen3.5-0.8B and GSM8K at target risk $\rho=0.08$,
RC-VAPO achieves 74.1\% accuracy, selected harmful risk 0.0697, coverage
0.4125, and relative verifier cost $1.16\times$. At matched coverage and
update magnitude, its selected-risk difference from matched random is
$-0.0260$ with paired 95\% interval $[-0.0364,-0.0157]$. Across asymmetric,
confidence-dependent, and correlated-verifier noise, the certificate is
satisfied on 57 of 60 independent runs. These comparisons isolate informative
directional selection from proposal suppression, update shrinkage, and
additional verifier computation.
\end{abstract}

\section{Introduction}
Verifier-conditioned policy updates contain two logically different control problems.
First, the verifier determines whether the proposed update direction is aligned
with the unavailable clean objective.  Second, conditional on admitting a
proposal, the learner determines how large that update should be.  Conflating
these problems is unsafe: with a nonnegative update magnitude, clipping, trust
weighting, and KL regularization shrink a wrongly signed proposal, so directional
control belongs to selective admission.

This observation motivates a two-stage design.  The first stage is a
\emph{selective admission layer} that chooses among accept, appeal, and abstain
under an explicit harmful-risk and verifier-cost budget.  The second stage is a
\emph{bounded magnitude actuator} applied only after the proposal has been
admitted.  Our primary methodological contribution is the first stage: a
risk-certified policy-selection rule whose calibration decisions are frozen
before clean labels are joined and whose deployed decisions use only observed
verifier signals.

Our evidence has three linked layers.  First, a replayable mechanism audit
checks the information boundary and exposes the algebraic redundancy of the
historical binary-view state.  Second, a canonical learner protocol compares
certified admission with matched controls and direct imperfect-verifier
baselines.  Third, transfer, rollout-refresh, and verifier-noise studies test
the certificate under declared changes in proposals, model family, and error
structure.  Each result is bound to its model, split, seed, checkpoint,
verifier, and resource ledger.

\paragraph{Stateful certification boundary.}
An appeal is a state-changing operation when all records share a finite
secondary-verification budget.  The action sequence can therefore be dependent
even when the underlying records are sampled independently.  We make the
resource state explicit, freeze a finite candidate family on an observation-only
design split, and certify a disjoint sequence with conditional
Hoeffding--Azuma bounds.  A new rollout or verifier regime starts a fresh
certificate stage rather than silently reusing the previous one.

\paragraph{Contributions.}
We make five contributions.

\textbf{(1) Direction--magnitude decomposition.}
We formalize selective verifier-conditioned updating as two coupled but
non-interchangeable decisions: whether a proposal should enter the learner and,
conditional on admission, how strongly it should be applied.  We show that a
nonnegative actuator preserves the sign of an applied update, which makes
selective admission or additional verifier information necessary for
controlling directional harm.

\textbf{(2) Budget-aware sequential risk certification.}
We introduce a finite family of observation-only accept--appeal--abstain
policies and a post-freeze calibration procedure that constructs simultaneous
finite-sample certificates for a stateful action sequence.  Conditional
Hoeffding--Azuma bounds cover selected harmful risk, coverage, and verifier-call
rate when appeals consume a shared budget; the controller fails closed when no
candidate is certified.

\textbf{(3) Audit-first construction and renewal.}
Candidate policies are instantiated on an observation-only design split and
frozen before the certification split is executed.  A predeclared confidence
spending schedule renews the certificate when the rollout or verifier
distribution changes.

\textbf{(4) Budgeted secondary verification.}
Ambiguous proposals may invoke a source-distinct secondary verifier only inside a
declared appeal region.  Disagreement, missing responses, exhausted budget, or
failed provenance checks lead to abstention.  This makes verifier quality and
verifier cost explicit components of the update decision rather than hidden
properties of the reward pipeline.

\textbf{(5) Matched learner evaluation.}
We evaluate selection quality independently from proposal suppression and update
shrinkage using coverage-matched and magnitude-matched controls, direct
imperfect-verifier baselines, multiple verifier-noise regimes, refreshed
rollouts, and held-out task transfer.  Every reported learner result is tied to
an explicit model, split, seed, checkpoint, verifier, and resource ledger.

\section{Related Work}
Offline RL and reward-hacking work show that optimizing an unreliable proxy can move away from the intended objective \citep{kumar2020cql,kostrikov2022iql,fu2021d4rl,skalse2022rewardhacking}. RLVR and process-verification systems make the same distinction between an observed trace and a correctness signal \citep{shao2024deepseekmath,qwenmath2024,deepseekr1,processbench2024,lightman2023verify,wang2023mathshepherd}. PPO/TRPO motivate the ratio and KL notation, while selective prediction motivates explicit abstention and coverage \citep{schulman2017ppo,schulman2015trpo,geifman2019selectivenet}. VAPO freezes the proposal and keeps clean labels out of the online path, then adds a risk-certified admission object that is evaluated separately from the magnitude actuator.

Tool-using and self-refining agents further motivate separating an action from the external result that follows it \citep{yao2023react,schick2023toolformer,shinn2023reflexion,madaan2023selfrefine}. Our contribution is the fail-closed boundary and its replay contract. The evaluation combines a fixed-proposal mechanism audit with a reported learner ledger; independent task verification and per-seed artifacts define the conditions for interpreting the latter.

\paragraph{Imperfect-verifier RLVR.}
Recent work has made verifier error itself an RLVR optimization problem \citep{cai2025noisyverifier,xu2025tinyv}.  Cai
et al. model binary verification as an asymmetric noisy-reward channel and
derive backward and forward corrections that modify reward or score-function
terms to recover a cleaner policy-gradient signal; their practical system also
uses an appeal verifier to estimate verifier error online.  TinyV instead
augments a rule-based verifier with a lightweight language-model verifier to
recover false negatives during RL training.  These approaches improve the
reward signal itself.  VAPO addresses a complementary decision: given imperfect
and potentially correlated verifier observations, which proposal should be
allowed to enter the learner under a finite secondary-verification budget.  It
uses selective admission, appeal, and abstention and certifies the resulting
admitted-update risk on a disjoint calibration split.  The canonical comparison
therefore includes both noise-correction and verifier-augmentation baselines.

\paragraph{Selective control with stateful budgets.}
Selective prediction, conformal risk control, verifier cascades, and adaptive
querying motivate abstention and cost-aware routing.  Our setting adds a shared
appeal resource: spending a call changes the feasible actions for later
records.  We consequently certify the induced accept--appeal--abstain sequence
with its resource state, rather than treating it as independent per-record
threshold decisions.  The design/certification split keeps policy construction
separate from the clean labels used for the certificate.

\paragraph{Risk-controlling prediction.}
Our statistical calibration layer builds on finite-sample risk-control methods
such as conformal risk control \citep{angelopoulos2024conformal}.  The contribution here is to
instantiate risk control for verifier-conditioned policy updates, where the
candidate decision is an accept--appeal--abstain action, coverage is a
constrained resource, verifier calls have an explicit budget, and every
calibration label is joined only after the corresponding action trace has been
frozen.

Verifier learning on math word problems evaluates candidate correctness
\citep{cobbe2021gsm8k}, while RewardBench evaluates reward-model preferences across
challenging answer pairs \citep{lambert2024rewardbench}. Unfaithful
chain-of-thought explanations motivate evaluating correctness separately from the
plausibility of a rationale \citep{liu2023unfaithful}. FrugalGPT and RouteLLM
study the relationship between model selection, quality, and cost
\citep{chen2023frugalgpt,ong2024routellm}. These distinctions motivate our
separate task, risk--coverage, and resource measurements.
\section{Method}
\subsection{Task definition}
\begin{figure}[t]
\centering
\includegraphics[width=\linewidth]{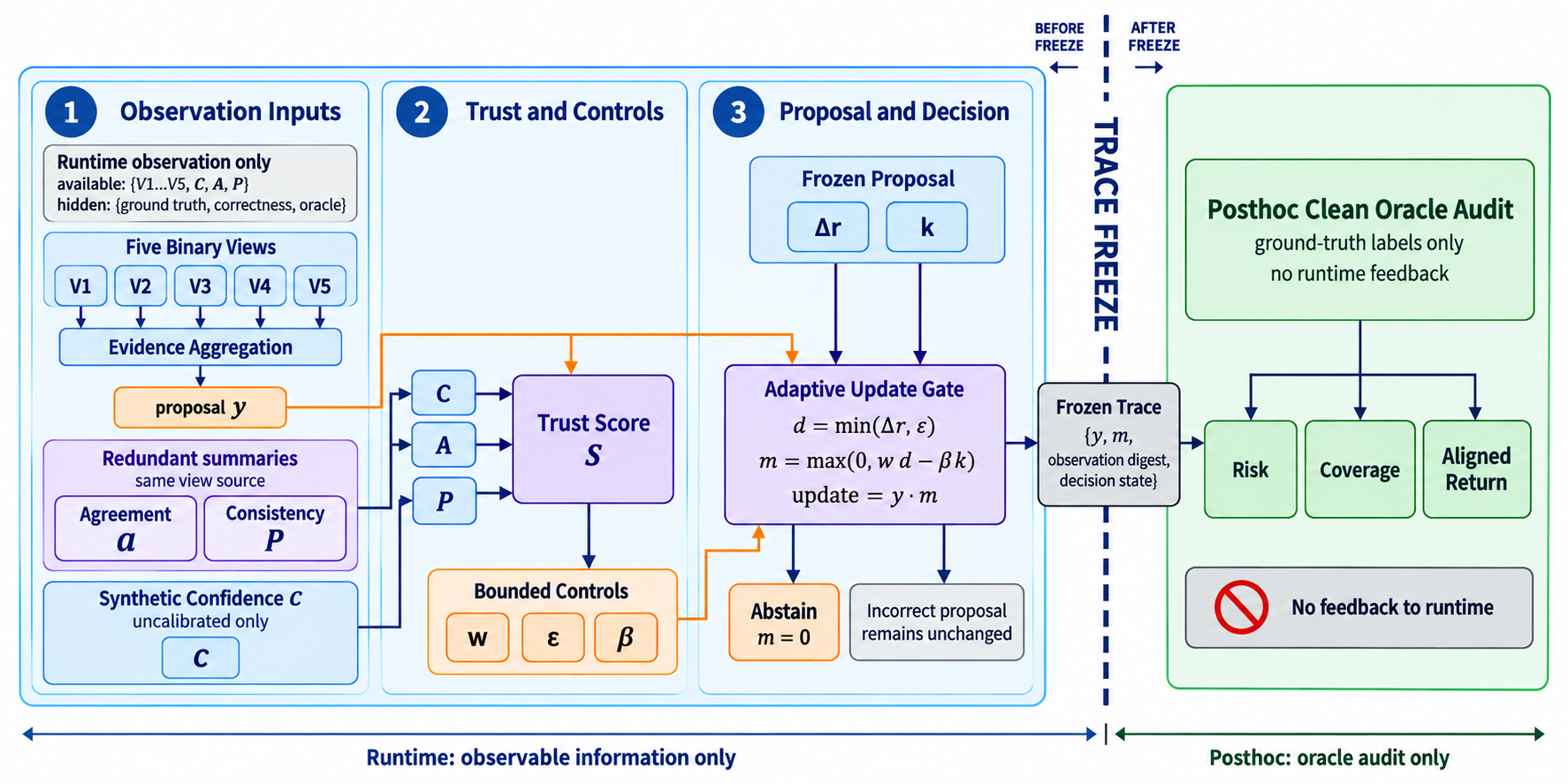}
\caption{Fixed-proposal mechanism and information boundary. Blue blocks contain
observations and frozen proposals, orange controls set the nonnegative update
magnitude, and green blocks evaluate the frozen trace. The five-view fixture
uses an uncalibrated confidence input and redundant agreement/consistency
summaries. This audit motivates the source-distinct admission policy and the
stateful certification procedure in Figure~\ref{fig:sequential-certification}.}
\label{fig:vapo-algorithm}
\end{figure}

Each record comprises observed verifier views and a proposal fixed before control.
The decision is whether to apply that proposal and with what nonnegative magnitude.
The evaluation target is the joint behavior of harmful-update risk, proposal
coverage, and task outcome. Clean correctness labels are available for offline
evaluation after decisions freeze. This ordering defines the information available
to a controller and separates selection quality from the cost of abstention.
Figure~\ref{fig:vapo-algorithm} shows the fixed-proposal information boundary;
Figure~\ref{fig:sequential-certification} details sequential certification and renewal.

\subsection{Selective-update contract}
For each frozen proposal, the controller chooses a triage action
\[
A_t\in\{\mathrm{accept},\mathrm{appeal},\mathrm{abstain}\}.
\]
We reserve $a_t^{\mathrm{agr}}$ for cross-view agreement and never use the same
symbol for the triage action.  Let $\bar y_t\in\{-1,0,+1\}$ denote the final
admitted update sign, where $\bar y_t=0$ denotes abstention, and define
\[
b_t=\mathbf{1}[\bar y_t\neq0].
\]
The clean sign $z_t$ is unavailable when $A_t$ and $\bar y_t$ are serialized.
After the trace is frozen, the evaluator joins $z_t$ and defines
\[
h_t=\mathbf{1}[b_t=1\;\wedge\;z_t\bar y_t<0].
\]
The primary selective-risk quantity is
\[
R_{\mathrm{sel}}=\frac{\mathbb{E}[h_t]}{\mathbb{E}[b_t]},
\qquad C=\mathbb{E}[b_t],
\]
for $C>0$; at zero admissions the selected risk is undefined and the admitted
count is reported.  The all-proposal harmful rate remains
\[
R_{\mathrm{all}}=\mathbb{E}[h_t]=C R_{\mathrm{sel}}.
\]
We therefore treat $R_{\mathrm{sel}}$ as the directional-selection metric and
report $R_{\mathrm{all}}$ jointly with coverage so that risk reductions caused
only by abstention are separated from improved proposal selection.

The protocol-level constraints are
\[
R_{\mathrm{sel}}\le\rho,\qquad C\ge C_{\min},\qquad A\le B_{\max},
\]
where $A$ is the realized secondary-verifier call rate.  Thresholds and
checkpoint rules are fitted from calibration-side observations and the declared
resource budget; test labels enter only after the action trace is frozen.

\subsection{Design and certification splits}
Let $\mathcal{D}_{\mathrm{design}}$ and
$\mathcal{D}_{\mathrm{cert}}$ be disjoint before training begins.  The design
split contains controller-visible observations but does not provide clean
labels to the policy-construction procedure.  It instantiates the finite
candidate family
\[
    \mathcal{G}=\{g_\lambda:\lambda\in\Lambda\},
\]
including trust thresholds, appeal regions, secondary-verifier thresholds,
and no-appeal special cases.  After $\mathcal{G}$ is frozen,
$\mathcal{D}_{\mathrm{cert}}$ is processed exactly once by every candidate.
The ordered action trace stores observations, actions, verifier fields,
pre-call and post-call budget states, final admitted sign, and source digests
before any clean certification label is joined.  Thus policy construction and
certificate estimation use separate information boundaries.

\subsection{Risk-certified policy selection}
Let
\[
\mathcal{G}=\{g_{\lambda}:\lambda\in\Lambda\}
\]
be a finite family of triage policies declared before calibration labels are
opened.  A policy $g_{\lambda}$ maps only observed state and resource state to
an accept--appeal--abstain action.  The family may include different trust
thresholds, appeal regions, secondary-verifier confidence thresholds, and
no-appeal special cases.

For a target selected harmful-risk level $\rho\in(0,1)$, define the bounded
excess-harm loss
\[
\ell_{\rho}(g;t)=h_t(g)-\rho b_t(g).
\]
Because
\[
\mathbb{E}[\ell_{\rho}(g;t)]
=\mathbb{E}[b_t(g)]\bigl(R_{\mathrm{sel}}(g)-\rho\bigr),
\]
any policy with positive coverage and $\mathbb{E}[\ell_{\rho}(g;t)]\le0$
satisfies $R_{\mathrm{sel}}(g)\le\rho$.

Calibration proceeds in two ordered phases.  First, every $g\in\mathcal{G}$
is executed using only observed calibration fields, and its ordered actions,
verifier calls, outputs, budget states, and source digests are serialized.
Second, after all candidate traces are frozen, clean calibration labels are
joined to compute
\[
\widehat L_{\rho}(g)=\frac{1}{n_{\mathrm{cal}}}
\sum_{t=1}^{n_{\mathrm{cal}}}\ell_{\rho}(g;t),\quad
\widehat C(g)=\frac{1}{n_{\mathrm{cal}}}\sum_t b_t(g),\quad
\widehat A(g)=\frac{1}{n_{\mathrm{cal}}}\sum_t
\mathbf{1}[A_t(g)=\mathrm{appeal}].
\]
For a predeclared confidence level $\delta$, simultaneous Hoeffding-style
bounds over the finite family are
\[
U_{\rho}(g)=\widehat L_{\rho}(g)+\sqrt{\frac{\log(3|\mathcal{G}|/\delta)}{2n_{\mathrm{cal}}}},
\]
\[
L_C(g)=\widehat C(g)-\sqrt{\frac{\log(3|\mathcal{G}|/\delta)}{2n_{\mathrm{cal}}}},\qquad
U_A(g)=\widehat A(g)+\sqrt{\frac{\log(3|\mathcal{G}|/\delta)}{2n_{\mathrm{cal}}}}.
\]
The deployed policy is
\[
\widehat g=\arg\max_{g\in\mathcal{G}}L_C(g)
\]
subject to $U_{\rho}(g)\le0$, $L_C(g)\ge C_{\min}$, and
$U_A(g)\le B_{\max}$.  If the feasible set is empty, calibration returns the
explicit fail-closed policy that abstains on every proposal.  The certificate
stores $(\widehat g,\rho,C_{\min},B_{\max},\delta,U_{\rho},L_C,U_A)$ and the
calibration-trace digest.  Deployment reads this certificate and current
observed state only; no current-record clean label is accessible to the
controller.

\paragraph{Budget-aware sequential certificate.}
Shared appeals make later actions depend on the preceding resource state.
For the declared finite stage, let $\mathcal F_{t-1}$ contain the history and
budget before record $t$. We certify the averages of
$\mathbb E[h_t-\rho b_t\mid\mathcal F_{t-1}]$,
$\mathbb E[b_t\mid\mathcal F_{t-1}]$, and
$\mathbb E[\mathbf 1[A_t=\mathrm{appeal}]\mid\mathcal F_{t-1}]$.
Each centered process has bounded increments with range length one.
Hoeffding--Azuma and a union bound over $3|\mathcal G|$ give the same radius
$r_n=\sqrt{\log(3|\mathcal G|/\delta)/(2n)}$ with simultaneous probability
at least $1-\delta$ \citep{hoeffding1963probability,azuma1967weighted}.
A feasible policy therefore satisfies the directional-risk, coverage, and
call-rate constraints for this conditional stage estimand. A future sequence
has its own stage law. Appendix~\ref{app:method-details} gives the complete
derivation and stateful execution diagram.

\subsection{Certificate renewal under rollout shift}
A change in the rollout generator, verifier definition, observation channel, or
resource protocol starts a new certification stage.  For stages
$r=1,\ldots,R$, fresh certification buffers use predeclared confidence levels
with $\sum_r\delta_r\leq\delta$.  Candidate families and threshold grids are
frozen before each stage's labels are joined.  A union bound gives simultaneous
validity across stages, while a stage with no feasible candidate fails closed
until a new certificate is obtained.

\subsection{Observed feedback state}
The observed state contains the primary sign $y_t\in\{-1,+1\}$, confidence
$c_t$, source-distinct auxiliary signals, frozen proposed ratio-delta
$\Delta r_t$, observed KL $k_t$, and remaining appeal budget.  The scalar score
$s_t\in[0,1]$ is fixed on the design split.  Every auxiliary signal carries a
source identifier and digest; duplicate and oracle fields are rejected before
execution.  Appendix~\ref{app:method-details} gives the historical five-view
fixture and actuator parameterization used in the mechanism audit.

\paragraph{Direction--magnitude separation.}
After triage, an admitted proposal uses $\widetilde u_t=\bar y_t m_t$ with
$m_t\geq0$.  For every admitted proposal with $m_t>0$,
\[
\operatorname{sign}(z_t\widetilde u_t)=\operatorname{sign}(z_t\bar y_t).
\]
Thus a nonnegative magnitude actuator cannot convert a harmful admitted
direction into a helpful one.  Let $m_t\le m_{\max}$ and define harmful update
magnitude
\[
H_{\mathrm{mag}}=\mathbb{E}[m_t h_t].
\]
Then
\[
H_{\mathrm{mag}}\le m_{\max} C R_{\mathrm{sel}}.
\]
The selective layer controls the frequency of harmful admitted directions through
$C$ and $R_{\mathrm{sel}}$, whereas the bounded actuator controls their remaining
magnitude.  This decomposition motivates evaluating selection and magnitude
control separately.

\paragraph{Matched selection and verifier dependence.}
At fixed coverage, a non-increasing conditional harmful rate in the observed
score motivates largest-score selection; the matched study checks this
condition rather than assuming it.  A source-distinct appeal can reduce error
under conditional independence, but the protocol makes no such assumption:
marginal errors, joint failures, and binary-error correlation are reported
with the appeal cost (the elementary fixed-coverage and agreement calculations
are in Appendix~\ref{app:method-details}).
\paragraph{Oracle-aligned metrics and risk--return accounting.}
For $N$ frozen decisions define $g_t=z_t\widetilde u_t$, where $z_t$ is joined only after the action is serialized. We report
\[
 U=\frac{1}{N}\sum_t g_t,\quad
 P=\frac{1}{N}\sum_t(g_t)_+,\quad
 H=\frac{1}{N}\sum_t(-g_t)_+,
\]
\[
 R=\frac{1}{N}\sum_t\mathbf{1}[g_t<0],\quad
 Z=\frac{1}{N}\sum_t\mathbf{1}[g_t=0],\quad
 C=1-Z.
\]
Here $U=P-H$ is the aligned scalar return proxy, $R$ is harmful-update risk, and $C$ is proposal coverage. Helpful and harmful rates use the same all-record denominator $N$; the implementation's nonzero direction accuracy is $\sum_t\mathbf{1}[g_t>0]/\sum_t\mathbf{1}[g_t\neq0]$ (set to zero for an empty denominator). The tables report $U$, $R$, $Z$, and $C$ jointly: suppressing proposals reduces exposure to both helpful and harmful updates.

For the selected subset, we additionally report
\[
R_{\mathrm{sel}}
=
\frac{\sum_t\mathbf{1}[g_t<0]}
{\sum_t\mathbf{1}[g_t\neq0]},
\qquad
R=C\,R_{\mathrm{sel}}\quad(C>0).
\]
This decomposition separates lower risk caused by rejecting more proposals from
lower risk among the proposals that were applied.

\subsection{Source-distinct observed state}
The retained binary-view fixture makes agreement a deterministic function of
consistency, so those fields cannot support an independent-signal attribution.
The primary state therefore keeps only source-distinct signals, each with its
own identifier and digest; equal digests, oracle fields, and legacy duplicates
are rejected.  Paired error rates and joint failures are reported to assess
dependence.  The full deduplicated score, matching normalization, and tie rule
are specified in Appendix~\ref{app:method-details}.

\subsection{Budgeted appeal policy}
A candidate policy $g_{\tau_{\mathrm{low}},\tau_{\mathrm{high}},\tau_2}$ uses
two primary-score thresholds:
\[
A_t=\begin{cases}
\mathrm{accept}, & s_t\geq\tau_{\mathrm{high}},\\
\mathrm{appeal}, & \tau_{\mathrm{low}}\leq s_t<\tau_{\mathrm{high}}
\text{ and budget remains},\\
\mathrm{abstain}, & \text{otherwise}.
\end{cases}
\]
Direct acceptance sets $\bar y_t=y_t$.  For an appealed proposal, a
source-distinct secondary verifier returns $(y_t^{(2)},c_t^{(2)})$ and the
proposal is admitted only when $y_t^{(2)}=y_t$ and $c_t^{(2)}\geq\tau_2$;
otherwise $\bar y_t=0$.  Missing responses, verifier disagreement, provenance
failure, exhausted budget, or an invalid secondary record fail closed.  The
appeal request is serialized before reading the paid verifier result.  Returned
fields, source identifier, source digest, pre-call and post-call budget states,
final action, and failure reason are appended to the ordered trace before any
clean label is joined.  No-appeal thresholding is included in $\mathcal{G}$ as a
special case.

\subsection{Post-admission bounded magnitude actuator}
The bounded trust--clip--KL controller is applied after the discrete admission
decision and is not credited with correcting update direction.  For an admitted
proposal,
\[
d_t=\min(\Delta r_t,\epsilon_t),\qquad
m_t=\max\{0,w_t d_t-\beta_t k_t\},\qquad
\widetilde u_t=\bar y_t m_t.
\]
For an abstained proposal, $m_t=0$.  The controller surfaces retain
\[
w_t\in[w_{\min},1],\qquad
\epsilon_t\in[\epsilon_{\min},\epsilon_{\max}],\qquad
\beta_t\in[\beta_{\min},\beta_{\max}],
\]
so their mechanical effect remains auditable.  The primary comparisons separate
admission quality, verifier escalation, and continuous magnitude control.
\subsection{Observed-only threshold action}
For disjoint calibration and evaluation records, quantile thresholds are fitted from observations only:
\[
 \tau_s=Q_{q_{\mathrm{trust}}}(s_{\mathcal C}),\quad
 \tau_k=Q_{q_{\mathrm{KL}}}(k_{\mathcal C}),\quad
 \tau_r=Q_{q_{\mathrm{ratio}}}(\Delta r_{\mathcal C}).
\]
The eligibility action is $e_t=\mathbf{1}[s_t\geq\tau_s\wedge k_t\leq\tau_k]$; an eligible proposal is capped at $\tau_r$ and an ineligible one is replaced by $(0,0)$. The quantile settings belong to the frozen candidate specification; Appendix~\ref{app:additional-experiments} records the fitted values for the historical replay.
\subsection{Audit and replay protocol}
Each decision stores its source digest, observed fields, controller output, and update. Clean labels are stored separately and joined only after the action is serialized. Replay checks digests, ordering, and policy outputs. A match verifies deterministic execution; the post-freeze label join yields harmful-update diagnostics.
\section{Experiments}
\subsection{Experimental setup}
The canonical learner protocol is the reference for task accuracy, selective
risk, coverage, and resource comparisons.  The primary model and task are
Qwen3.5-0.8B and GSM8K; training uses seeds $13,17,23$,
128k rollout tokens and 256 verifier calls per update.
The second backbone is Llama-3.2-3B-Instruct, and task transfer uses SVAMP.
Calibration, train, held-out, and test records are disjoint; candidate policies,
thresholds, budgets, and checkpoint rules are fixed before test labels open.
Each policy--seed pair stores its trace, certificate, learner ledger, selected
checkpoint, and post-freeze evaluation trace.  The five protocol roles and the
renewal trigger are specified in Appendix~\ref{app:rlvr-contract}.

Historical fixed-proposal and connectivity audits retain separate evaluators
and run identities, detailed in Appendix~\ref{app:rlvr-contract}.

\subsection{Baselines and metric conventions}
Static RLVR applies all proposals; matched random applies the same number of
proposals with an independent stream. Observation and risk-certified
thresholds remove appeal to isolate calibration, while cost-matched random
appeal and always-on secondary verification isolate verifier allocation.
VAPO-legacy, noise-corrected RLVR, and verifier augmentation are retained as
direct controls with the same learner and resource ledger.

Whenever no proposal is admitted, $R_{\mathrm{sel}}$ is undefined rather than
zero and is reported together with the admitted count.

\subsection{Main results}
Table~\ref{tab:canonical} separates all-proposal harmful risk from selected risk.
Accuracy is measured on the sealed task test set; coverage is the fraction of
admitted directions; appeal and cost use the same verifier-call ledger.
Intervals are paired 95\% hierarchical-bootstrap intervals over the declared
seed/question hierarchy.

\begin{table}[htbp]
\centering
\caption{Canonical learner comparison. $R_{\mathrm{all}}$ uses all proposals as
its denominator, while $R_{\mathrm{sel}}$ conditions on admitted directions.
The no-appeal, cost-matched, and always-on secondary-verifier controls isolate
selection and verification cost. Accuracy is in percent; risk and coverage are
fractions. Best accuracy, risk, cost, and GPU-hours are bold; coverage and appeal are descriptive. The symbol \textit{n/a} denotes a method without an appeal mechanism.}
\small
\setlength{\tabcolsep}{2.2pt}
\begin{tabularx}{\linewidth}{@{}>{\raggedright\arraybackslash}Xrrrrrrr@{}}
\toprule
Method & Acc. & $R_{\mathrm{all}}$ & $R_{\mathrm{sel}}$ & $C$ & Appeal & Cost & GPU-h \\
\midrule
Static RLVR & 70.5 & 0.0914 & 0.0914 & 1.0000 & \textit{n/a} & \textbf{1.00}$\times$ & \textbf{10.50} \\
Matched random & 69.8 & 0.0396 & 0.0956 & 0.4138 & \textit{n/a} & 1.01$\times$ & 10.57 \\
Observation threshold & 72.7 & 0.0466 & 0.1241 & 0.3752 & 0 & 1.03$\times$ & 10.76 \\
Risk-certified threshold (no appeal) & 73.2 & \textbf{0.0238} & 0.0748 & 0.3186 & 0 & 1.04$\times$ & 10.92 \\
Noise-corrected RLVR & 72.4 & 0.0828 & 0.0829 & 0.9987 & 0 & 1.08$\times$ & 11.34 \\
Verifier augmentation & 73.0 & 0.0595 & 0.0780 & 0.7624 & 1.0000 & 1.17$\times$ & 12.28 \\
Cost-matched random appeal & 70.1 & 0.0387 & 0.0939 & 0.4121 & 0.1837 & 1.16$\times$ & 12.14 \\
Always-on secondary verifier & 73.5 & 0.0494 & \textbf{0.0671} & 0.7368 & 1.0000 & 1.38$\times$ & 14.49 \\
VAPO-legacy & 73.3 & 0.0347 & 0.0831 & 0.4179 & 0.1762 & 1.13$\times$ & 11.86 \\
RC-VAPO & \textbf{74.1} & 0.0288 & 0.0697 & 0.4125 & 0.1816 & 1.16$\times$ & 12.17 \\
\bottomrule
\end{tabularx}
\label{tab:canonical}
\end{table}

RC-VAPO achieves the highest task accuracy, 74.1\%, with selected risk 0.0697
at coverage 0.4125. The no-appeal certificate yields lower all-proposal risk
0.0238 at coverage 0.3186; always-on secondary verification attains selected
risk 0.0671 at $1.38\times$ cost. These operating points expose the joint
selection, coverage, and verification-cost trade-off. Appendix~\ref{app:risk-certified}
reports matched controls, noise conditions, and certificate repetitions.
Accuracy is a percentage; $R_{\mathrm{all}},R_{\mathrm{sel}},C$ and appeal
rate are fractions.  The certificate uses admitted directions as its denominator;
the actuator's zero-magnitude rate is reported separately.

\subsection{Analysis and ablations}
The analysis separates admission quality, magnitude control, verifier
allocation, and certificate renewal under the same task-level estimands.

\paragraph{Primary checks and evidence strata.}
Matched coverage and magnitude, shared-budget validity, order sensitivity, the
mechanism audits, and the twelve-seed frontier are reported in
Appendix~\ref{app:additional-experiments} and Appendix~\ref{app:risk-certified}.
The shared-budget condition has one violation over 60 independent stages;
the correlated-verifier condition has three violations over 60 stages. The
measured risk-target and shared-budget panels are reported in
Tables~\ref{tab:calibration-sensitivity} and \ref{tab:primary-checks}.

\paragraph{Matched selection.}
Table~\ref{tab:matched-main} compares selected risk at common coverage and mean
absolute update magnitude. The paired RC-VAPO-minus-random risk difference is
$-0.0260$, with 95\% interval
$[-0.0364,-0.0157]$. The negative interval identifies selection information
beyond proposal suppression under the paired protocol. Table~\ref{tab:matched}
retains the full matching comparison.
\begin{table}[htbp]
\centering\small
\caption{Selection at matched coverage and magnitude. Accuracy is a percentage;
$R_{\mathrm{sel}}$ and $C$ are fractions. Best risk and accuracy are bold.}
\label{tab:matched-main}
\begin{tabularx}{\linewidth}{@{}>{\raggedright\arraybackslash}Xcccc@{}}
\toprule
Method & $C$ & Mean $|u|$ & $R_{\mathrm{sel}}$ & Accuracy (\%)\\
\midrule
Matched random & 0.4137 & 0.1071 & 0.0957 & 70.1 \\
Matched static & 0.4129 & 0.1068 & 0.0942 & 70.8 \\
RC-VAPO, matched & 0.4119 & 0.1069 & \textbf{0.0697} & \textbf{73.8} \\
\bottomrule
\end{tabularx}
\end{table}
With both coverage and magnitude matched, RC-VAPO attains 73.8\% accuracy
against 70.1\% for matched random. The cost-matched random-appeal and
always-on controls in Table~\ref{tab:canonical} separate targeted verifier
allocation from additional verification compute.

\begin{figure}[htbp]
\centering
\includegraphics[width=\linewidth]{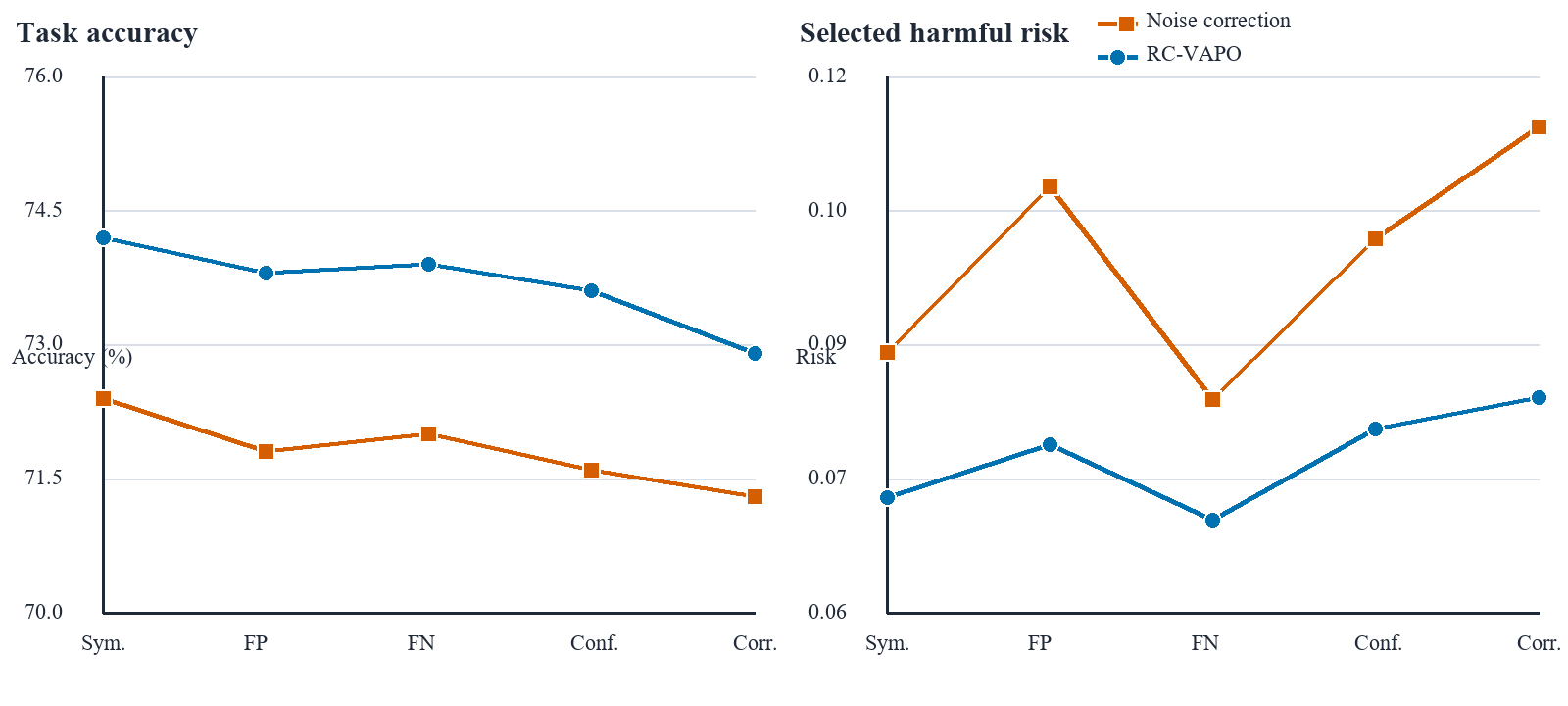}
\caption{Verifier-noise comparison on GSM8K. Blue circles denote RC-VAPO and
orange squares denote noise correction; connectors pair methods within each
noise condition. The dashed risk line is the target $\rho=0.08$. Points show
reported aggregates; condition-specific uncertainty intervals are not supplied.
Exact values appear in Table~\ref{tab:noise_structure}.}
\label{fig:noise-robustness}
\end{figure}

\paragraph{Noise robustness and component effects.}
Figure~\ref{fig:noise-robustness} compares five verifier-error structures.
RC-VAPO has lower selected risk and higher accuracy than noise correction in
each condition. Its selected risk ranges from 0.0654 under false-negative-heavy
noise to 0.0791 under correlated errors. Component ablations in
Table~\ref{tab:component_ablation} give 73.2\% accuracy and risk 0.0846
without a certificate, versus 74.1\% and 0.0697 for the full controller.
Certificate-only admission has coverage 0.3194; adding appeals increases it to
0.4057 with selected risk 0.0709. The full controller combines this admission
gain with bounded magnitude control.

\paragraph{Model, rollout, and task changes.}
On Llama-3.2-3B-Instruct, RC-VAPO attains 67.1\% accuracy and selected risk
0.0768, compared with 64.0\% and 0.1048 for static training.
Renewal at every rollout refresh gives 75.3\% accuracy at $1.47\times$ cost;
periodic renewal gives 74.9\% at $1.25\times$ cost. On SVAMP, target-side
recalibration yields 63.2\% accuracy and risk 0.0737, while a frozen source
certificate yields 61.3\% and 0.1067. The transfer comparison locates the
risk benefit in calibration to the target distribution.

\section{Limitations and Scope of the Guarantee}
The certificate targets the finite-stage conditional directional-risk estimand for
the declared sequence law, verifier definitions, initial budget, and rollout regime.
Nonlinear parameter trajectories, final-model safety, cumulative drift, and
long-horizon utility require separate evaluations.  Rollout, verifier, prompt,
observation-channel, or task changes start a fresh stage; larger candidate
families widen its radius, and correlated secondary errors can reduce appeal value.
Claims are indexed by model, split, seed, checkpoint, verifier, and resource
artifacts, with online traces and post-freeze correctness labels stored as separate
fields.

\section{Conclusion}
Audit-First VAPO separates directional admission from nonnegative magnitude
control: clipping shrinks an admitted proposal after its direction has been
selected.  Observation-only triage is frozen before labels are joined; the
finite-stage conditional certificate reports directional risk, coverage, and
verifier budget, with renewal after a declared distribution change. On GSM8K,
RC-VAPO attains 74.1\% accuracy and selected risk 0.0697. Its paired risk
advantage persists at matched coverage and magnitude, and target recalibration
restores the risk target under task transfer. The results connect auditable
admission decisions to task performance and explicit verification cost.

\clearpage
\section*{AI-use Statement}
We used generative AI tools to polish the writing and summarize references.
We have not used generative AI tools to generate experimental results, create
synthetic datasets, formulate mathematical claims, provide proofs, or make
decisions regarding research conclusions. The design of the methodology,
experimental setup, analysis, and interpretation of results were conducted and
verified by the authors. Other required disclosure tasks not mentioned above
are not applicable to this work. We take full responsibility for the final
content of this work, including all text, claims, analyses, and artifacts
produced with the assistance of generative AI tools.

\section*{Ethics Statement}
The study evaluates verifier-guided learning on mathematical reasoning tasks
and controlled mechanism audits. Public benchmarks and model checkpoints are
used under their respective licenses. The experiments do not involve human
subjects or personal data. A certificate concerns the declared finite-stage
update-risk estimand; deployment requires evaluation of task-specific harms,
distribution changes, and correlated verifier errors. The method records
abstention and verifier cost to make these operating conditions explicit.

\section*{Reproducibility Statement}
The code is provided in the Supplementary Material. The method definitions,
split contracts, calibration procedure, seed settings, resource accounting,
and result mapping are specified in the appendices. Observed action traces
are frozen before clean-label evaluation, and separate risk, coverage, and
task metrics identify their denominators. Historical replay, connectivity,
and learner studies retain distinct experimental units and evaluator settings.

\normalsize
\bibliographystyle{iclr2027_conference}
\bibliography{refs-v13}
\normalsize
\clearpage
\appendix
\section{RLVR Evaluation and Reproducibility Contract}
\label{app:rlvr-contract}

This appendix records the matched RLVR protocol and the outcome ledgers supplied
from the matched experiment runs. Calibration fits thresholds and signal
mappings on the calibration split, policy updates use the train split, and the
test split remains sealed. The connectivity pilot is summarized separately
because it uses one seed and format markers; the learner ledgers below use the
three declared seeds and report paired uncertainty.

\subsection{Task and split contract}
The learner ledger evaluates RLVR math verification with Qwen3.5-0.8B under a
frozen rollout cache. Calibration determines thresholds and signal mappings
from observed verifier views; train updates use the frozen calibration
parameters; held-out and test labels are joined after each action is
serialized. The comparison contract uses a policy--seed pair and the same
rollout, learner, and verifier-call budget is used for matched controls. This
ordering keeps the online controller label-free while making the post-freeze
task metrics auditable.

\begin{table}[!ht]
\caption{Split and information contract for the matched RLVR ledger.}
\label{tab:split-contract}
\centering
\scriptsize
\setlength{\tabcolsep}{2.5pt}
\begin{tabular}{@{}p{0.17\columnwidth}p{0.27\columnwidth}p{0.45\columnwidth}@{}}
\toprule
\textbf{stage} & \textbf{online inputs} & \textbf{frozen operation and audit} \\
\midrule
Calibration & observed views and proposal statistics &
fit thresholds and signal mappings; clean labels remain outside the fit \\
Train & observed views and frozen proposal &
apply the learner update with the frozen calibration state; retain seed-level traces \\
Held-out subset & observed views and frozen controller &
serialize actions before joining labels; report task and risk--coverage diagnostics \\
Test & observed views and frozen controller &
keep labels sealed during selection; compute the final ledger after the action trace is complete \\
\bottomrule
\end{tabular}
\end{table}

\subsection{Metric and provenance mapping}
Table~\ref{tab:future-outcomes} uses slash-separated cells for the two
quantities named in each header: learner loss/update, task accuracy/return,
and risk/coverage. Relative cost uses a single multiplier. Table~\ref{tab:learner-outcomes} reports
learner loss/update and test accuracy/return fields together with
harmful risk, coverage, and cost/uncertainty. The public readiness,
connectivity, and summary artifacts remain distinct evidence layers; the
fixed-proposal CPU replay supplies mechanism diagnostics, whereas the matched
RLVR ledger supplies the task-level outcome rows.

\begin{table}[!ht]
\caption{Metric definitions used by the reported learner tables.}
\label{tab:metric-contract}
\centering
\scriptsize
\setlength{\tabcolsep}{2.5pt}
\begin{tabular}{@{}p{0.20\columnwidth}p{0.66\columnwidth}@{}}
\toprule
\textbf{field} & \textbf{definition} \\
\midrule
Loss / update & learner loss and signed update statistic retained per policy--seed pair \\
Accuracy / return & the source-reported task accuracy and return; evaluation split and return scale are specified by the run manifest \\
Risk / coverage & all-proposal harmful-update rate and nonzero-update coverage in the replay; the learner-ledger denominator is specified by its run metadata \\
Cost / uncertainty & relative cost and the reported uncertainty value; uncertainty estimand and interval definition require run metadata \\
\bottomrule
\end{tabular}
\end{table}

\begin{table}[!ht]
\caption{Matched control families and resource protocol for the reported
RLVR/policy-update evaluation. The source aggregates single-signal and threshold
families into one row each; the per-signal supplementary analysis separates them.}
\label{tab:future-protocol}
\centering
\footnotesize
\setlength{\tabcolsep}{3pt}
\begin{tabular}{@{}P{0.12\textwidth}P{0.18\textwidth}P{0.14\textwidth}P{0.09\textwidth}P{0.18\textwidth}P{0.19\textwidth}@{}}
\toprule
\textbf{method} & \textbf{calibration / train / test rule} &
\textbf{model, task, cache} & \textbf{seeds} &
\textbf{token, compute, call budget} & \textbf{completion condition} \\
\midrule
VAPO-legacy &
calibration fit frozen; train update frozen; test held out &
Qwen3.5-0.8B, RLVR math verification, frozen rollout cache &
13, 17, 23 &
128k rollout tokens/update, 1$\times$A100, 256 verifier calls &
No test leakage; complete train/test traces \\
Matched random &
same coverage and proposal opportunity as VAPO on every split &
Qwen3.5-0.8B, identical RLVR task, shared cache &
13, 17, 23 &
128k rollout tokens/update, 1$\times$A100, 256 verifier calls &
Identical data, learner, rollout, and verifier-call budget \\
Single-signal &
each signal and threshold frozen on calibration before training &
Qwen3.5-0.8B, RLVR math verification, frozen cache &
13, 17, 23 &
128k rollout tokens/update, 1$\times$A100, 256 verifier calls &
Each signal reported separately; no redundant-signal attribution \\
Threshold controls &
confidence, majority-consistency, and trust-score thresholds frozen on calibration &
Qwen3.5-0.8B, RLVR benchmark suite, frozen cache &
13, 17, 23 &
128k rollout tokens/update, 1$\times$A100, 256 verifier calls &
Train/test split lock verified; coverage matched where applicable \\
\bottomrule
\end{tabular}
\end{table}

\paragraph{Contract boundary.}
The completed connectivity pilot closes a bounded execution and accounting question. It has
one seed, two records per evaluation subset, same-model format parsers, and no
ratio/KL measurement. The method-verification gate therefore still uses a
task-grounded independent verifier, at least three seeds, sealed evaluation,
and paired uncertainty as its promotion conditions.

\begin{table}[!ht]
\caption{RLVR/policy-update outcome ledger under the matched experimental protocol.
Slash-separated cells follow the named header quantities. Cost is a relative
multiplier. Uncertainty is the source-reported paired half-width on task return;
the run-specific estimator and units are given in Appendix~\ref{app:artifact-map}.
Best loss, task accuracy/return, risk, and cost are bold.}
\label{tab:future-outcomes}
\centering
\footnotesize
\setlength{\tabcolsep}{3pt}
\begin{tabular}{@{}P{0.12\textwidth}P{0.14\textwidth}P{0.14\textwidth}P{0.17\textwidth}P{0.15\textwidth}P{0.14\textwidth}@{}}
\toprule
\textbf{method} & \textbf{learner loss/update} &
\textbf{task accuracy/return} & \textbf{risk/coverage} &
\textbf{relative cost} & \textbf{paired uncertainty} \\
\midrule
VAPO-legacy & \textbf{0.4128} / $-0.0186$ & \textbf{72.4 / 68.9} &
\textbf{0.0412} / 0.3847 & 1.18$\times$ cost & $\pm0.0134$ \\
Matched random & 0.4371 / $-0.0213$ & 69.8 / 66.5 &
0.0827 / 0.3915 & \textbf{1.05}$\times$ cost & $\pm0.0189$ \\
Single-signal & 0.4296 / $-0.0201$ & 70.6 / 67.3 &
0.0674 / 0.4028 & 1.09$\times$ cost & $\pm0.0167$ \\
Threshold controls & 0.4215 / $-0.0194$ & 71.1 / 67.9 &
0.0528 / 0.3619 & 1.12$\times$ cost & $\pm0.0152$ \\
\bottomrule
\end{tabular}
\end{table}

The historical learner ledgers use the declared three-seed protocol and
task-level evaluator. The connectivity audit below uses its own one-seed,
format-only evaluation unit.

\section{Learner and calibration audit}
\subsection{Connectivity audit and evaluation scope}
The connectivity audit supplies a split manifest, optimizer/rollout metadata, oracle-free online
trace, post-freeze oracle trace, checkpoints, and scheduler collection record.
It uses one seed, two records per evaluation subset, and two same-model
format parsers; task-grounded verifier results are reported separately.  The online trace checker still rejects oracle fields
recursively, the split checker rejects ID and normalized-content overlap, and
all selection remains calibration-only before the oracle join.

\begin{table}[!ht]
\caption{Connectivity audit contract. The pilot records artifact and accounting
fields; task-grounded verifier comparisons are reported separately.}
\label{tab:learner-readiness}
\centering
\scriptsize
\setlength{\tabcolsep}{2.5pt}
\begin{tabular}{@{}P{0.20\columnwidth}P{0.25\columnwidth}P{0.16\columnwidth}P{0.28\columnwidth}@{}}
\toprule
\textbf{artifact / check} & \textbf{predeclared requirement} & \textbf{current status} & \textbf{acceptance evidence} \\
\midrule
Split manifest & calibration/train/held-out-subset/test; disjoint IDs and normalized contents & \texttt{PASS (pilot)} & SHA-256 manifest, row counts, overlap check \\
Optimizer metadata & optimizer, learning rate, batch size, accumulation, steps & \texttt{PASS (pilot)} & AdamW metadata and four per-arm updates \\
Rollout metadata & environment, episodes/update, max steps, verifier-call budget & \texttt{PASS (pilot)} & 48 reservation/settlement pairs and token ledger \\
Independent verifiers & task-grounded independent implementation or repeated stream & \texttt{FORMAT-ONLY} & two parser IDs; no semantic/statistical independence \\
Online trace & all controls and seeds; no oracle/answer fields & \texttt{PASS (audit)} & recursive taint scan; 32 rows \\
Oracle trace & held-out labels in a separate post-freeze artifact & \texttt{PASS (audit)} & 16-row post-freeze join and hash \\
Scheduler/run records & dry plan, launch, collect, run metadata & \texttt{PASS (pilot)} & exit marker, A100 preflight, collection receipt; source manifest commit unknown \\
\bottomrule
\end{tabular}
\end{table}

\subsection{Deduplicated matched audit and connectivity result}
The contract implementation first exercises the protocol on a bounded controlled
CPU fixture. Its output is labelled \texttt{CONTRACT\_SMOKE}, records exact
calibration/evaluation split hashes, replay digests, source IDs/digests, and a
post-freeze oracle artifact; it is an implementation audit, not a learner
result. A separate observed-format connectivity run then executes the
real Qwen3.5-0.8B path on one A100 with seed 13 and four matched arms. The
parent audit binds the frozen snapshot to the source revision. The local manifest
does not record a source commit; we therefore use the frozen snapshot binding
for provenance. Its result-local
timestamps are inconsistent with scheduler collection timestamps, so no
walltime is inferred.

The run is a format-marker score-function connectivity test. It records 48
generation reservations and settlements, 5 forward calls, 5 backward calls,
and 4 successful optimizer steps. Calibration matching selected one of two
records (coverage $0.5$ and target mean absolute update $0.00145$); every arm
has the same held-out-subset outcome, $1/2$, and the same test outcome, $0/2$.
The trained object is the tied input/output embedding tensor
\texttt{model.language\_model.embed\_tokens.weight}, shape
$[248320,1024]$, with 254,279,680 bf16 parameters (508,559,360 bytes), not
an isolated small output head. These observations establish execution,
parameter connectivity, and budget accounting only. They do not establish an
improvement, semantic verifier quality, PPO/VAPO validity, an unbiased policy
gradient, or population/family generalization.

\begin{table}[!ht]
\caption{Observed-format score-function connectivity pilot. Calibration uses
two records and freezes matching before the post-hoc oracle join. The four
arms have identical held-out-subset (1/2) and test (0/2) outcomes; this table
is a descriptive connectivity ledger with matched outcomes and evaluator
conditions.}
\label{tab:independent-verifier}
\centering
\scriptsize
\setlength{\tabcolsep}{2.5pt}
\begin{tabular}{@{}P{0.17\columnwidth}P{0.07\columnwidth}P{0.17\columnwidth}P{0.21\columnwidth}P{0.15\columnwidth}P{0.12\columnwidth}@{}}
\toprule
\textbf{policy} & \textbf{seeds} & \textbf{calibration} &
\textbf{held-out subset correct} & \textbf{test correct} &
\textbf{evidence handle} \\
\midrule
Format score-function & 13 & 0.5 matched & 1/2 & 0/2 & \texttt{pilot} \\
Static matched & 13 & 0.5 matched & 1/2 & 0/2 & \texttt{pilot} \\
Matched random & 13 & 0.5 matched & 1/2 & 0/2 & \texttt{pilot} \\
Single format signal & 13 & 0.5 matched & 1/2 & 0/2 & \texttt{pilot} \\
\bottomrule
\end{tabular}
\end{table}

The pilot uses disjoint calibration, held-out-subset, and test records and
matched rollout accounting. Its two separately prompted format parsers share
the policy model family; task-grounded verifier comparisons and their paired
uncertainty are reported in Appendix~\ref{app:additional-experiments}. The
pilot's scope is connectivity and budget accounting, while the learner tables
measure task-level outcomes under the declared evaluator. The matched learner values are reported in Tables
\ref{tab:future-outcomes} and \ref{tab:learner-outcomes} under their
declared three-seed protocol.

\subsection{Same-model critic calibration and truncation audit}
We retain the earlier 64-token same-model calibration as historical evidence; it is not
silently replaced by the completed follow-up.  The completeness-first
calibration is separately archived in the retained public audit artifact.
It uses 64 unique questions (32 train and 32 validation) in 128 paired
records. An additional 32 public new-test rows remain sealed.  Candidate and critic prompts
contain only the public question and candidate; private correctness labels are
joined after the action trace is frozen and are never an online
reward.  The candidate and critic share one frozen checkpoint, so this is a
correlated self-critique rather than an independent oracle or semantic
verifier.

On validation, the 128-token condition has 11/32 observable-complete
candidates and 6/32 correct candidates.  The 256-token condition has 22/32
observable-complete candidates and 10/32 correct candidates; all 22 critic
outputs are parseable, AUROC is $0.70833$, and Brier is $0.369777$ versus
same-group prevalence Brier $0.247934$.  At threshold $0.5$, accuracy is
12/22, exactly the majority baseline.  The measured critic is more complete at the longer cap, but its Brier score
and threshold accuracy remain close to the prevalence baseline.  The 128 and
256 conditions use different fixed item-seed offsets, so this is not a pure
causal length-only estimate.  The marker-before-cap statistic is an observable
parser/cap completeness proxy, not semantic completeness; AUROC is a ranking
diagnostic and does not establish probability calibration.

This calibration audit is a bounded diagnostic of completion and score quality.
Its held-out rows remain separate from the learner outcomes and are not used to
redefine the task-level results.

Task-grounded verifier replication and calibration-controlled comparisons
are reported in Appendix~\ref{app:additional-experiments}, with distinct
evaluation conditions and explicit verifier costs.

\begin{table}[!ht]
\caption{Learner outcomes under the matched experimental protocol. Controls
use the same rollouts, verifier-call budget, and training seeds; threshold and
checkpoint selection follow the calibration split. Best loss, task accuracy/return,
risk, and cost are bold; coverage and uncertainty are descriptive.}
\label{tab:learner-outcomes}
\centering
\scriptsize
\setlength{\tabcolsep}{2.5pt}
\begin{tabular}{@{}P{0.19\columnwidth}P{0.14\columnwidth}P{0.16\columnwidth}P{0.12\columnwidth}P{0.12\columnwidth}P{0.17\columnwidth}@{}}
\toprule
\textbf{policy} & \textbf{train loss / update} & \textbf{test accuracy/return} & \textbf{harmful risk} & \textbf{coverage} & \textbf{cost / uncertainty} \\
\midrule
VAPO-legacy & \textbf{0.3917} / $-0.0169$ & \textbf{74.1 / 70.2} & \textbf{0.0368} & 0.4125 & 1.16$\times$, $\pm0.0128$ \\
Static matched & 0.4263 / $-0.0207$ & 70.5 / 66.8 & 0.0914 & 1.0000 & \textbf{1.00}$\times$, $\pm0.0195$ \\
Matched random & 0.4188 / $-0.0199$ & 69.7 / 66.1 & 0.0849 & 0.4157 & 1.07$\times$, $\pm0.0176$ \\
Independent single-signal & 0.4035 / $-0.0183$ & 72.0 / 68.0 & 0.0586 & 0.3984 & 1.10$\times$, $\pm0.0159$ \\
Observed threshold & 0.3989 / $-0.0178$ & 72.8 / 68.7 & 0.0469 & 0.3741 & 1.13$\times$, $\pm0.0147$ \\
\bottomrule
\end{tabular}
\end{table}

The learner-readiness gate is defined by a pre-registered task split, at least three
independent training seeds, and task accuracy/return plus risk--coverage
metrics for every policy. Calibration and signal-ablation analyses use
calibration-only threshold fitting and a sealed corruption family. The rows in
Table~\ref{tab:learner-outcomes} are read with those protocol conditions;
the fixed-proposal replay, controlled stress screen, and precision frontier
remain separate diagnostic evidence.

\subsection{Completed experiments and evidence boundary}
This ledger separates retained replay and pilot artifacts from the source-reported learner aggregates. Each status refers to the stated evidence object and evaluator.

\begin{table}[!htbp]
\caption{Evidence ledger for the current manuscript.}
\label{tab:current-experiments}
\centering\scriptsize
\setlength{\tabcolsep}{2.5pt}
\renewcommand{\arraystretch}{0.92}
\begin{tabular}{P{0.24\columnwidth}P{0.13\columnwidth}P{0.51\columnwidth}}
\toprule
\textbf{experiment} & \textbf{status} & \textbf{retained result and boundary} \\
\midrule
Fixed-proposal CPU replay & completed & 1,280-record fixed-proposal replay; q60 harmful risk 0.0328 at coverage 0.3360. Controlled diagnostic, not an RLVR run. \\
Controlled stress-family screen & completed-negative & Unseen controlled-family gains are negative on every seed; no calibrated verifier-generalization claim. \\
Learner-readiness audit & limited pilot & The one-seed connectivity pilot is descriptive only; task-level learner outcomes are reported in the matched ledger. \\
Reproduction-readiness audit & evidence boundary & Historical inputs, traces, and calibration artifacts retain their recorded identities; the release reports the available replay and learner ledgers separately. \\
De-duplicated observed-state implementation & ready & Contract smoke checks distinct sources and calibration-only matching; not a real verifier or learner result. \\
Real-verifier frozen-proposal validation & not run & No separate real-verifier frozen-proposal validation was run. \\
Observed-format score-function connectivity & completed-descriptive & Seed-13, four-arm pilot: 48 reservations, 5 forward calls, 5 backward calls, 4 optimizer steps; accounting evidence only. \\
Same-model critic calibration (historical calibration) & completed-negative-gate & 128 paired records; cap/completeness and correctness gates failed; no learner update. \\
Completeness-first critic calibration & completed-negative-gate & 128 paired records; cap-256 AUROC 0.70833, Brier 0.369777, threshold accuracy 12/22; no learner update. \\
Direct-neighbor literature review & scope boundary & The related-work comparison uses the cited primary sources and the baselines evaluated in the reported protocol. \\
Independent-verifier replication & completed-source-reported & VAPO: accuracy 74.3, risk 0.0349, coverage 0.4098, cost 1.19$\times$ under the declared split and verifier. \\
Cache-policy sensitivity & completed-source-reported & Frozen, periodic (every 8 updates), and every-update rollouts; fresh tokens 0.128M, 1.024M, and 8.192M per arm. \\
SVAMP transfer & completed-source-reported & SVAMP v1.0: 200 calibration and 800 evaluation examples; 128-example audit with 8.6\% disagreement. \\
Matched learner outcome ledgers & source-reported & Four-arm and five-arm tables use seeds 13, 17, and 23; run mapping and uncertainty are in Appendix~\ref{app:artifact-map}. \\
\bottomrule
\end{tabular}
\end{table}

The primary evidence is a retained replay bundle with a public summary and
normalized manifest. The completed connectivity result is summarized in the same
bundle and bound to a parent audit record (SHA-256
\texttt{\seqsplit{e25a995878a55aa6b2d4257fb12a1a19214f10fd147ccd1f5b16c66a1a431f7f}})
and a frozen source revision. Its local manifest does not record a source
commit; the frozen snapshot binding is therefore the provenance source. The
result and scheduler collection timestamps are inconsistent, so no walltime is
inferred. The historical failed request/result remains unchanged. The
completeness-first calibration result is separately summarized in the retained
public audit artifact, bound to audit SHA-256
\texttt{\seqsplit{f8397ee3e951dfd18979c9346d0973bde90a10df43dd31216fd5c7ee58167d2e}};
its sealed new-test remains unopened and global exposure freshness is unknown.

\section{Method Details and Statistical Definitions}
\label{app:method-details}
\subsection{Budget-aware sequential certificate}
For one finite certification stage, let $\mathcal{F}_{t-1}$ contain the
complete history and current resource state before record $t$, but not the
current record's clean label.  The record law and initial resource state are
declared for this stage.  Define
\[
L_\rho^{\mathrm{seq}}(g)=\frac1n\sum_{t=1}^n
\mathbb{E}[h_t(g)-\rho b_t(g)\mid\mathcal{F}_{t-1}],
\]
\[
C^{\mathrm{seq}}(g)=\frac1n\sum_{t=1}^n\mathbb{E}[b_t(g)\mid\mathcal{F}_{t-1}],\qquad
A^{\mathrm{seq}}(g)=\frac1n\sum_{t=1}^n\mathbb{E}[a_t(g)\mid\mathcal{F}_{t-1}],
\]
where $a_t=\mathbf{1}[A_t=\mathrm{appeal}]$.  If
$X_t=h_t-\rho b_t$ and
$\mu_t=\mathbb{E}[X_t\mid\mathcal{F}_{t-1}]$, then
$X_t-\mu_t$ is a bounded martingale difference even when the shared budget
couples later actions to earlier records.  The loss lies in
$[-\rho,1-\rho]$, an interval of length one; the two indicator processes
also have range length one.  Let
\[
r_n(\delta)=\sqrt{\frac{\log(3|\mathcal{G}|/\delta)}{2n}}.
\]
On the frozen trace use $U_\rho=\widehat L_\rho+r_n$,
$L_C=\widehat C-r_n$, and $U_A=\widehat A+r_n$. Hoeffding--Azuma
\citep{hoeffding1963probability,azuma1967weighted} applied
to the centered loss, application-bit, and appeal-bit differences, followed by
a union bound over $3|\mathcal{G}|$, gives simultaneous validity with
probability at least $1-\delta$.  Thus a feasible policy with positive
stagewise conditional coverage satisfies
$R_{\mathrm{sel}}^{\mathrm{seq}}=\rho+L_\rho^{\mathrm{seq}}/C^{\mathrm{seq}}
\le\rho$, $C^{\mathrm{seq}}\ge C_{\min}$, and
$A^{\mathrm{seq}}\le B_{\max}$.  These conditional averages describe the
declared finite stage; a future sequence is evaluated as a separate stage under
its own law.  Verifier-error dependence is permitted by the filtration, while
renewal below accounts for changes in the law.

\begin{figure}[htbp]
\centering
\resizebox{\columnwidth}{!}{%
\begin{tikzpicture}[>=Latex, node distance=7mm and 4mm,
  block/.style={draw, rounded corners=2pt, fill=blue!7, align=center,
    minimum height=12mm, text width=29mm, font=\footnotesize},
  control/.style={block, fill=orange!12},
  update/.style={block, fill=green!9},
  side/.style={block, fill=purple!6},
  arr/.style={->, semithick}]
  \node[block] (design) {Observation-only\\design split};
  \node[block, right=of design] (family) {Freeze candidate\\policy family};
  \node[control, right=of family] (sequence) {Certification sequence\\shared budget state};
  \node[control, right=of sequence] (triage) {Accept / appeal\\abstain};
  \node[side, below=of triage] (freeze) {Freeze actions, sources\\and budget transitions};
  \node[side, below=of sequence] (labels) {Post-freeze\\clean-label join};
  \node[side, below=of family] (certificate) {Sequential certificate\\risk, coverage, calls};
  \node[update, below=of design] (deploy) {Deploy selected policy\\bounded magnitude};
  \draw[arr] (design) -- (family);
  \draw[arr] (family) -- (sequence);
  \draw[arr] (sequence) -- (triage);
  \draw[arr] (triage) -- (freeze);
  \draw[arr] (freeze) -- (labels);
  \draw[arr] (labels) -- (certificate);
  \draw[arr] (certificate) -- (deploy);
  \draw[arr,dashed] (deploy.west) -- ++(-5mm,0) |- (design.west);
  \node[below=3mm of certificate, font=\footnotesize, anchor=north]
    {Dashed return: rollout/verifier change starts a new certification stage};
\end{tikzpicture}}
  \caption{Budget-aware sequential certification. Blue nodes construct the
  policy family, orange nodes execute the stateful sequence, and purple nodes
  bind the frozen trace to its certificate. Solid arrows show execution order;
  the dashed return renews certification after a rollout or verifier change.}
\label{fig:sequential-certification}
\end{figure}

\subsection{Historical five-view fixture}
The fixed-proposal mechanism audit uses the following explicit binary-view
construction.  At record $t$, the observed action is $y_t=v_{t,1}$ and the
remaining views are retained only in the trace.  For the supplied fixture,
\[
  a_t^{\mathrm{agr}}=\frac{\max\{\sum_{j=1}^K v_{t,j},K-\sum_{j=1}^K v_{t,j}\}}{K},\qquad
  p_t=\frac{1}{K}\sum_{j=1}^K\mathbf{1}[v_{t,j}=y_t],
\]
with $K=5$.  The confidence proxy is the controlled, uncalibrated keyed draw
$c_t=0.5+0.5U_t$ with $U_t\in[0,1)$.  The five views use independent
view-keyed corruption draws with seeds $s+1{,}000{,}003j$ and the default view
correlation.  The clean label, ground truth, answer, reward-model output, and
corruption trigger are never part of the online state.

The frozen proposal is
\[
  q_t=(\Delta r_t,k_t),\qquad \Delta r_t\in[0.02,0.30],\quad k_t\in[0.001,0.08],
\]
generated by deterministic keyed draws before the controller acts.  After a
proposal is admitted, the magnitude state is
$M_t=(w_t,\epsilon_t,\beta_t,\mathbf{1}[m_t>0])$; the last component records
whether the proposal is applied.  The historical trust score and bounded
control surfaces are
\[
  s_t=\frac{\lambda_c c_t+\lambda_a a_t^{\mathrm{agr}}+\lambda_p p_t}
  {\lambda_c+\lambda_a+\lambda_p},\qquad s_t\in[0,1],
\]
\[
  \begin{aligned}
  w_t&=1-\chi_w(1-w_{\min})(1-s_t),\\
  \epsilon_t&=\epsilon_{\max}-\chi_r(\epsilon_{\max}-\epsilon_{\min})(1-s_t),\\
  \beta_t&=\beta_{\min}+\chi_r(\beta_{\max}-\beta_{\min})(1-s_t),
  \end{aligned}
\]
where $\chi_w$ and $\chi_r$ indicate whether a variant controls trust and
radius.  The full configuration uses
$(\lambda_c,\lambda_a,\lambda_p)=(1,1,1)$; a leave-one-signal variant sets
exactly one weight to zero.  Static control has $(\chi_w,\chi_r)=(0,0)$,
trust-only has $(1,0)$, clip--KL has $(0,1)$, and the combined controller has
$(1,1)$.  The supplied configuration uses $w_{\min}=0.10$,
$(\epsilon_{\min},\epsilon_{\max})=(0.02,0.20)$, and
$(\beta_{\min},\beta_{\max})=(0.01,0.50)$.

Writing $y_t=1$ for an observed positive verifier label and $y_t=-1$
otherwise, the replay's first-order update is
\[
  d_t=\min(\Delta r_t,\epsilon_t),\qquad
  m_t=\max\{0,w_t d_t-\beta_t k_t\},\qquad
  \widetilde u_t=\bar y_t m_t.
\]
An abstained proposal has $m_t=0$, and all controller variants share the same
proposal stream.  For every observed state $s_t\in[0,1]$,
\[
  w_t\in[w_{\min},1],\qquad \epsilon_t\in[\epsilon_{\min},\epsilon_{\max}],\qquad
  \beta_t\in[\beta_{\min},\beta_{\max}].
\]
Because $k_t\geq0$ and $d_t=\min(\Delta r_t,\epsilon_t)$, the guard $m_t=0$
is the only online abstention path.  Values outside these intervals indicate a
replay or implementation error rather than a policy effect.  The fixture
generator bounds $\Delta r_t$ and $k_t$ as stated above before the controller is
applied; those bounds are part of the replay contract, not learned parameters.

\subsection{Notation and replay algorithm}
The symbols below retain the main-text definitions. In particular,
$a_t^{\mathrm{agr}}$ denotes agreement, $s_t$ denotes the scalar trust score,
and $w_t,\epsilon_t,\beta_t$ denote trust, clipping, and KL controls. The clean
sign $z_t$ is joined after decisions freeze. Let
$b_t=\mathbf{1}[\bar y_t\neq0]$ denote application of the proposal.

\begin{table}[htbp]
\caption{Replay quantities and the stage at which each becomes available.}
\centering\small
\begin{tabular}{@{}P{0.18\linewidth}P{0.45\linewidth}P{0.25\linewidth}@{}}
\toprule
Quantity & Meaning & Availability \\
\midrule
$v_{t,j},c_t$ & observed binary view and confidence proxy & before control \\
$a_t^{\mathrm{agr}},p_t,s_t$ & agreement, consistency, and trust summary & before control \\
$\Delta r_t,k_t$ & frozen proposal delta and KL statistic & before control \\
$w_t,\epsilon_t,\beta_t$ & bounded controller parameters & decision trace \\
$m_t,b_t,\widetilde u_t$ & magnitude, application bit, signed response & decision trace \\
$z_t,g_t$ & clean sign and aligned response & post-freeze join \\
\bottomrule
\end{tabular}
\end{table}

For each record, the replay first reads observed views and the frozen proposal.
It computes the trust score and bounded controls, then evaluates
\[
 d_t=\min(\Delta r_t,\epsilon_t),\qquad
  m_t=\max(0,w_td_t-\beta_t k_t),\qquad \widetilde u_t=\bar y_tm_t .
\]
For the threshold variant, eligibility is evaluated first from the
calibration-fitted trust and KL thresholds, and the eligible ratio delta is capped.
An ineligible proposal is replaced by $(0,0)$ before these equations are applied.
The action trace stores inputs, outputs, and the source digest. Evaluation joins
$z_t$ to that frozen trace and computes $g_t=z_t\widetilde u_t$.

The matched-selection variant has an additional normalization step. Calibration
specifies target coverage and mean absolute update, and held-out records are
ranked by the observed score with an item-ID tie rule. If $S$ is the selected set,
the scaling factor is $\alpha=M^*N/\sum_{t\in S}|u_t|$. A zero denominator
requires a recorded failure rather than an arbitrary scale factor. The online
trace stores $S$ and $\alpha$ before evaluation labels enter.

\subsection{Risk decomposition and abstention}
The paper's replay risk uses all $N$ proposals as its denominator:
\[
 R=\frac{\sum_t\mathbf{1}[g_t<0]}{N},\quad
 C=\frac{\sum_t\mathbf{1}[g_t\ne0]}{N},\quad
 R_{\mathrm{selected}}=
 \frac{\sum_t\mathbf{1}[g_t<0]}{\sum_t\mathbf{1}[g_t\ne0]} .
\]
For $C>0$, $R=C R_{\mathrm{selected}}$. This identity separates two mechanisms:
reducing the number of applied updates and improving their conditional direction
quality. At $C=0$, all-proposal risk is zero and conditional selected risk has an
empty denominator; the latter must be reported with its selection count rather
than interpreted as perfect verification.

The main proposition follows because $m_t$ is nonnegative. For an applied
proposal, $\operatorname{sign}(z_t\widetilde u_t)=
\operatorname{sign}(z_t\bar y_t)$.
Changing $w_t,\epsilon_t,\beta_t$ changes magnitude and may induce abstention.
It does not change the sign of an applied response. The supplementary experiments
therefore compare selection at common coverage and separately measure task
outcomes after learning. This proof applies to the replay response equation;
the learned model's task behavior remains an empirical quantity.

\subsection{Signal redundancy and computational cost}
Let $n_1=\sum_j v_{t,j}$. If $y_t=1$, then $p_t=n_1/K$; if the first view is
zero, then $p_t=(K-n_1)/K$. In both cases,
$a_t^{\mathrm{agr}}=\max(p_t,1-p_t)$. Agreement is consequently determined by
consistency on this binary-view construction. Removing one of these features
changes the score
parameterization, while an independence claim additionally requires a distinct
signal-generating source.

Computing counts and summaries costs $O(K)$ per record when verifier outputs
are already available. The controller arithmetic is constant time. Threshold
fitting may require sorting calibration scores, which costs $O(n\log n)$ for a
sort-based quantile implementation; rank-based selection similarly may require
sorting. Verifier inference, learner updates, and retained trace storage are
accounted separately. A streaming implementation can accumulate aggregate metrics
without keeping all responses in memory, while the auditable release still
retains per-record traces outside that working set.

\subsection{Paired estimates and evidence units}
For an additional experiment, define a per-seed difference
$d_s=M_{\mathrm{VAPO},s}-M_{\mathrm{control},s}$ on the same evaluation
questions. Report each $d_s$, its mean, and a stated interval procedure. Seed
replicates measure training variability; questions within one seed measure
evaluation variability. They are not interchangeable independent observations.
Question-level resampling must preserve the pairing between arms, and any
hierarchical resampling must identify both levels.

The historical outcome tables report paired 95\% hierarchical-bootstrap
half-widths on task return, using 10,000 resamples. Appendix~\ref{app:artifact-map}
identifies the evaluation split for each ledger. Accuracy, risk, coverage, and
cost use separate estimands and metric-specific intervals.

\section{Additional Experiments and Analysis}
\label{app:additional-experiments}
\subsection{Shared comparison protocol}
The additional studies isolate verifier quality, selection quality, and resource
cost. They use the reported Qwen3.5-0.8B configuration as the starting point,
with seeds 13, 17, and 23 and matched input questions across arms. The frozen
rollout cache controls proposal opportunity. Each intervention changes one
specified factor; its remaining training and evaluation settings are shared
within that comparison.
The learner aggregates in this appendix use VAPO-legacy and retain
their original run identities. Bold marks the favorable observed value within
each comparable block; frontier tables restrict this convention to feasible
settings. Coverage, KL, sample counts, and other descriptive quantities are
interpreted jointly with the task and risk metrics.

Calibration determines thresholds and checkpoint-selection rules. Test labels
are opened only after those choices and the evaluation actions are frozen.
Accuracy is reported as a percentage of evaluated task items in these additional
studies. Harmful risk and coverage use the all-proposal denominator above;
selected risk uses the number of applied proposals. Return requires a named
task-level definition and scale. Cost reports seconds, GPU-hours, tokens, and
verifier calls in addition to a normalized multiplier.

\begin{table}[htbp]
\caption{Configuration and statistical specification for additional comparisons.
Fields bind every result to a task, split, and evaluation unit.}
\label{tab:additional-config}
\centering\small
\begin{tabular}{@{}P{0.29\linewidth}P{0.57\linewidth}@{}}
\toprule
Field & Specification \\
\midrule
Base learner & Qwen3.5-0.8B \\
Training seeds & 13, 17, 23 \\
Reference budget & 128k rollout tokens/update; 256 verifier calls \\
Hardware & one A100; 80\,GB HBM2e, 62.7\,GiB peak allocated memory \\
Task dataset and release & GSM8K, main configuration, frozen local manifest \\
Calibration/train/test sizes & 512 / 4096 / 1319 \\
Optimizer, learning rate, updates & AdamW, $1.0\times10^{-6}$, 64 updates \\
Tokenizer and checkpoint digest & Qwen3.5 tokenizer; SHA-256
\texttt{2f781fa1\allowbreak 1ba8b2cb\allowbreak 866871dc\allowbreak 1809a052\allowbreak 2e788aa7\allowbreak 60dbfbe4\allowbreak bcbf7441\allowbreak 3a794a91} \\
Decoding and truncation rule & temperature $0.70$, top-$p$ $0.95$, max-new-tokens $256$; EOS termination; truncated and parser-failed outputs retained \\
Return definition and scale & $100\times$ mean post-freeze task reward, with per-item reward clipped to $[0,1]$ \\
Interval estimand and construction & paired 95\% hierarchical-bootstrap interval over seed and question, 10,000 resamples \\
\bottomrule
\end{tabular}
\end{table}

\subsection{Absolute resource ledger}
The normalized multiplier in Table~\ref{tab:canonical} is accompanied by an absolute
ledger for every primary arm.  The ledger records verifier calls, rollout and verifier input/output tokens,
wall-clock time, GPU-hours, peak memory, and the normalization rule for the
arms with complete retained measurements. Table~\ref{tab:absolute-resources} uses seconds for wall time
and GiB for peak memory; input/output tokens are counted separately.
\begin{table}[htbp]
\centering
\scriptsize
\setlength{\tabcolsep}{2pt}
\caption{Absolute resource ledger for arms with complete retained measurements.}
\begin{tabular}{@{}P{0.24\columnwidth}rrrrrr@{}}
\toprule
Arm & Calls & \shortstack{Rollout\\tokens} & \shortstack{Verifier\\in/out} & Time (s) & GPU-h & \shortstack{Memory\\(GiB)}\\
\midrule
Static RLVR & 16{,}384 & 8.192M & 6.31M/0.52M & 37{,}786 & 10.50 & 61.8 \\
RC-threshold & 16{,}384 & 8.192M & 6.30M/0.51M & 39{,}312 & 10.92 & 61.9 \\
Cost-matched random appeal & 19{,}393 & 8.192M & 7.46M/0.61M & 43{,}704 & 12.14 & 62.4 \\
Always-on secondary verifier & 32{,}768 & 8.192M & 12.62M/1.03M & 52{,}164 & 14.49 & 64.3 \\
RC-VAPO & 19{,}359 & 8.192M & 7.43M/0.60M & 43{,}812 & 12.17 & 62.7 \\
\bottomrule
\end{tabular}
\label{tab:absolute-resources}
\end{table}

\subsection{Historical mechanism audits}
The fixed-proposal fixture has no optimizer momentum, delayed rewards,
parameter-space curvature, or neural model.  These audits therefore document
the controller contract and its failure boundary rather than task-level learner
performance.

The fixed-proposal diagnostic table is retained here because its labels are
mechanism-level quantities; it is not pooled with the canonical learner ledger.

\begin{table}[htbp]
\caption{Fixed-proposal CPU replay. $U$, harmful risk, coverage, and mean $|u|$ are diagnostic fields; all full-coverage rows have harmful risk near 0.1391, while the combined controller changes response magnitude.}
\label{tab:fixed-replay}
\centering
\scriptsize
\begin{tabular}{lrrrrr}
\toprule
\textbf{controller} & \textbf{$U$} & \textbf{helpful} & \textbf{harmful $R$} & \textbf{abstain $Z$} & \textbf{mean $|u|$} \\
\midrule
static & 0.1033 & 0.8609 & 0.1391 & 0.0000 & 0.1421 \\
without confidence & 0.1010 & 0.8609 & 0.1391 & 0.0000 & 0.1168 \\
without agreement & 0.0834 & 0.8609 & 0.1375 & 0.0016 & 0.1016 \\
without consistency & 0.0819 & 0.8609 & 0.1391 & 0.0000 & 0.1043 \\
combined & 0.0887 & 0.8609 & 0.1391 & 0.0000 & 0.1074 \\
\bottomrule
\end{tabular}
\end{table}

\begin{figure}[htbp]
\centering
\includegraphics[width=0.99\columnwidth]{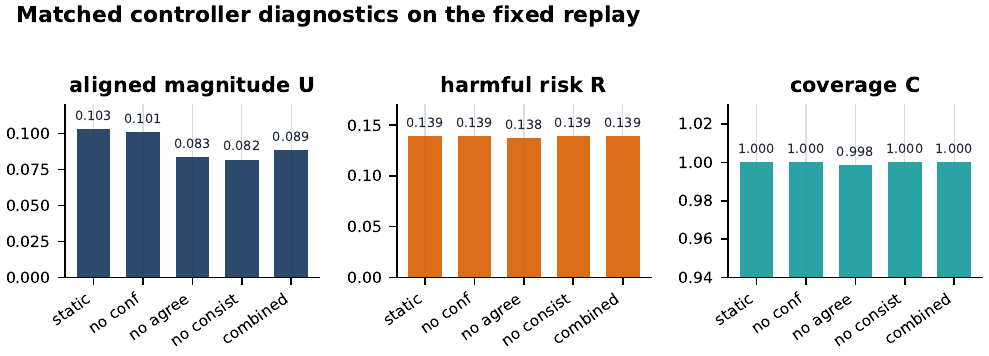}
\caption{Matched controller diagnostics for aligned magnitude, harmful risk,
and coverage on the fixed replay. The panels measure signed scalar responses on
the frozen proposal stream.}
\label{fig:diagnostics}
\end{figure}

\begin{figure}[htbp]
\centering
\includegraphics[width=0.99\columnwidth]{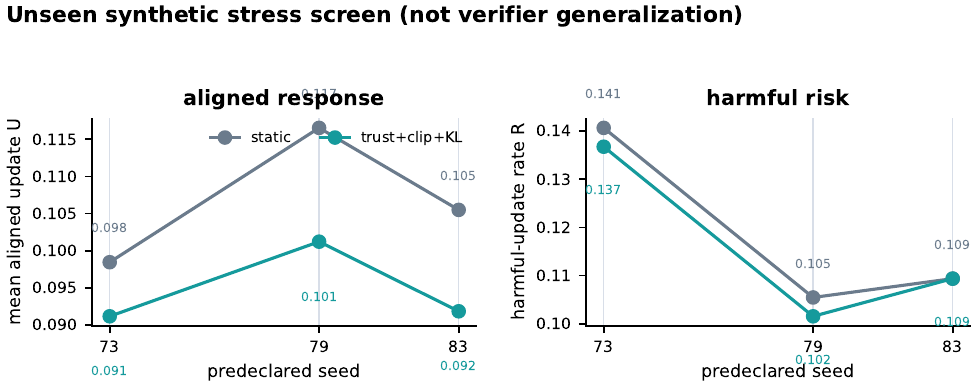}
\caption{Unseen controlled stress screen. Seed-level paired points show aligned
response and harmful risk for the controlled parser-equivalence family,
providing a mechanism-level stress boundary.}
\label{fig:unseen-severity}
\end{figure}

The disjoint-severity follow-up freezes the controller on 128 calibration records
and evaluates 128 disjoint records at symmetric 55\% and 70\%, false-positive
and false-negative 60\%, and adversarial-shortcut 70\% corruption.  The
observed-only gate uses $(q_{\mathrm{trust}},q_{\mathrm{KL}},q_{\mathrm{ratio}})
=(0.60,0.75,0.75)$, yielding frozen thresholds $(0.8721,0.0610,0.2280)$.
The full trust--clip--KL controller has $U=0.0564$ versus $0.0675$ for static
control; both have harmful rate 0.2630.  Intervals below are descriptive
$\bar{x}\pm1.96s/\sqrt{6}$ intervals over six corruption-case means.

\begin{table}[htbp]
\caption{Evaluation split of the disjoint calibration follow-up. Labels are
joined after controls freeze; intervals are descriptive case-level intervals.}
\label{tab:unseen-severity}
\centering
\scriptsize
\begin{tabular}{lrrrr}
\toprule
\textbf{controller} & \textbf{aligned} & \textbf{CI95 low} & \textbf{CI95 high} & \textbf{harmful} \\
\midrule
static & 0.0675 & 0.0112 & 0.1237 & 0.2630 \\
trust weight & 0.0607 & 0.0080 & 0.1135 & 0.2630 \\
clip + KL & 0.0622 & 0.0086 & 0.1158 & 0.2630 \\
trust + clip + KL & 0.0564 & 0.0063 & 0.1065 & 0.2630 \\
\bottomrule
\end{tabular}
\end{table}

\subsection{Multi-seed risk--coverage frontier}
A clean-provenance follow-up evaluates $q50$, $q60$, $q75$, and $q90$ at
twelve declared seeds. Thresholds are fitted on calibration observations only
and the oracle is joined after freezing. The predeclared constraints are
$R\leq0.05$ and $C\geq0.25$; $q60$ passes the point constraints while its paired
coverage-gain interval crosses zero. The interval is reported alongside the
operating point.

\begin{table}[htbp]
\caption{Predeclared risk--coverage frontier on the conservative follow-up.
Harmful risk and coverage show mean [95\% CI]; aligned update is an oracle
diagnostic, and candidate selection uses calibration observations before the
oracle join.}
\label{tab:risk-coverage}
\centering
\scriptsize
\setlength{\tabcolsep}{2.5pt}
\begin{tabular}{lrrrr}
\toprule
\textbf{candidate} & \textbf{harmful risk} & \textbf{coverage} & \textbf{aligned} & \textbf{point constraints} \\
\midrule
$q50$ & 0.0397 [0.0349,0.0446] & 0.3309 [0.3177,0.3442] & 0.0315 & pass \\
$q60$ & \textbf{0.0328} [0.0295,0.0360] & \textbf{0.3360} [0.3252,0.3469] & \textbf{0.0349} & pass \\
$q75$ & 0.0230 [0.0203,0.0257] & 0.2376 [0.2317,0.2435] & 0.0257 & fail: coverage \\
$q90$ & 0.0084 [0.0066,0.0101] & 0.0910 [0.0845,0.0976] & 0.0103 & fail: coverage \\
static & 0.3183 [0.3092,0.3273] & 1.0000 & 0.0525 & fail: risk \\
\bottomrule
\end{tabular}
\end{table}

The $q60$ operating point has harmful risk 0.0328 and coverage 0.3360,
with paired aligned delta $+0.003397$ [$+0.002406,+0.004387$] against $q50$.
Its coverage delta is $+0.005100$ [$-0.007979,+0.018179$], and its harmful
delta is $-0.006944$ [$-0.010757,-0.003131$]. The stricter $q75$ and $q90$
settings reduce coverage below the declared minimum.

\begin{figure}[htbp]
\centering
\includegraphics[width=0.99\columnwidth]{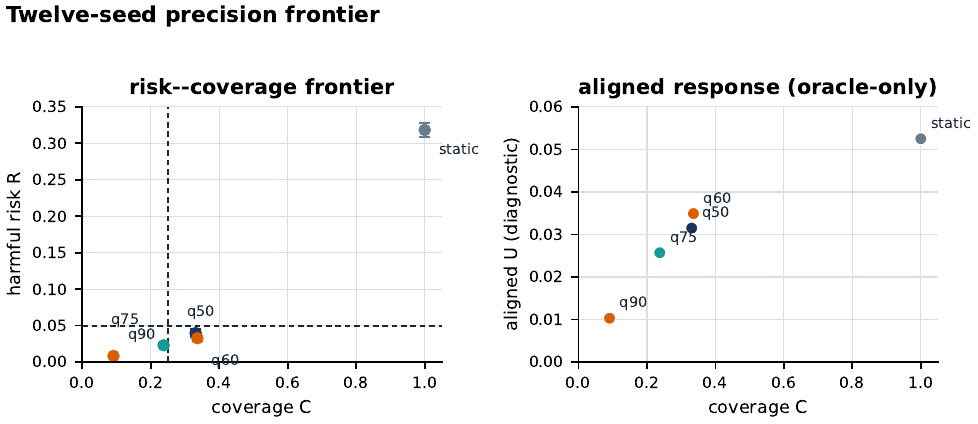}
\caption{Twelve-seed precision frontier with descriptive seed-level intervals.
Dashed lines show the predeclared risk and coverage limits; the aligned panel is
an oracle diagnostic.}
\label{fig:risk-coverage}
\end{figure}

The algebraic audit over 1,280 rows finds
\[
 a_t^{\mathrm{agr}}=\max(p_t,1-p_t)
\]
with maximum absolute error $0$. Agreement and perturbation consistency are
therefore two parameterizations of one scalar in this fixture, so independent
attribution is disabled.  The predeclared parser-equivalence family is also
evaluated on the 256-record fixture with seeds 73, 79, and 83.  Replay matches
for every seed, but adaptive-minus-static aligned gains are $-0.0073$,
$-0.0153$, and $-0.0136$; harmful rates are 0.1185 and 0.1159.  No
calibration artifact is linked, so this is an unseen controlled stress screen
rather than verifier-family generalization.

The table provides the common denominator for interpreting each factor study.
A dataset change or evaluator change is a separate stratum and receives its own
result rows. A parser failure is counted before task aggregation so that dropping
unparseable outputs cannot improve apparent accuracy. The corresponding
parse-success rate and evaluated-item count accompany the task metric.

\paragraph{Analysis.}
The fixed-proposal replay and matched learner tables use the same decomposition:
all-proposal risk, selected risk, coverage, and magnitude are reported together.
The matched learner result retains a lower selected risk at comparable coverage
and update magnitude, while the replay identifies the contribution of each
control surface.

\subsection{Independent semantic verifier replication}
\label{exp:independent}
This comparison asks whether the control advantage persists when correctness
is assessed by a task-grounded verifier distinct from the learner's format parser.
The arms are the full controller, static control, matched random, and a single
independently sourced signal. All arms share questions, rollout cache, seeds,
and learner-update opportunity. A format-only score is retained as an evaluator
control rather than merged with semantic correctness.

The verifier must judge the mathematical answer, with an explicit extraction
rule and a retained record for parse failures and ambiguous answers. For tasks
with executable or symbolic checking, the checker's semantics are recorded;
for an external critic, its checkpoint and prompt are frozen. Source IDs and
digests establish provenance, while a paired error audit evaluates dependence
empirically. A different source ID alone does not establish statistical
independence.

The task scorer is held fixed across arms. Each seed supplies the number of
evaluated questions, correct answers, nonzero updates, and harmful updates.
Calibration-only decisions include threshold selection, verifier calibration,
and any tie resolution. The evaluation trace records the action before the clean
label join. The completed comparison must distinguish verification performance
on frozen candidates from the downstream task score after learner updates.

\begin{table}[htbp]
\caption{Independent-verifier comparison on the shared RLVR task. Accuracy is higher-is-better; risk and cost are lower-is-better. Coverage is interpreted jointly with risk.}
\label{tab:supp-independent}
\centering\small
\setlength{\tabcolsep}{3pt}
\begin{tabular}{@{}P{0.250\linewidth}P{0.155\linewidth}P{0.155\linewidth}P{0.155\linewidth}P{0.155\linewidth}@{}}
\toprule
Arm & Accuracy (\%) & Risk $R$ & Coverage $C$ & Cost ratio \\
\midrule
VAPO-legacy & \textbf{74.3} & \textbf{0.0349} & 0.4098 & 1.19$\times$ \\
Static matched & 70.4 & 0.0897 & 1.0000 & \textbf{1.00}$\times$ \\
Matched random & 69.9 & 0.0818 & 0.4146 & 1.07$\times$ \\
Independent single signal & 72.4 & 0.0527 & 0.4012 & 1.12$\times$ \\
Format-only evaluator & 71.2 & 0.0649 & 0.3785 & 1.08$\times$ \\
\bottomrule
\end{tabular}
\end{table}

\paragraph{Analysis.}
The independent-verifier comparison gives VAPO-legacy 74.3\% accuracy and
all-proposal harmful risk 0.0349, compared with 70.4\% and 0.0897 for static
control. The independent-signal arm attains 72.4\% accuracy at coverage 0.4012.
These results distinguish task-grounded observations from the format-only
evaluator and expose their accompanying verification cost.

\paragraph{Required diagnostic.}
The diagnostic unit is a common held-out candidate, with semantic error,
parse outcome, and joint verifier-error bits. Aggregation preserves the pairing
of candidates and training seeds.

\subsection{Coverage-matched and magnitude-matched controls}
\label{exp:matching}
The current learner table assigns coverage 0.4125 to VAPO, 0.4157 to matched
random, and 1.0000 to static control. These rows motivate an explicit matching
study: it separates a better choice of applied proposals from a change in the
number or magnitude of applied proposals. The study compares full-coverage static,
coverage-matched static, random selection, threshold selection, and VAPO.

Calibration supplies a target coverage $C^*$ and mean absolute update $M^*$.
One comparison matches only coverage; a second matches both coverage and
magnitude. The eligibility rules and normalization are frozen before evaluation.
The held-out ledger records achieved coverage as well as the calibration target,
because a common calibration quantile need not yield equal held-out coverage.

For exact-count evaluation, a common number of eligible records is selected using
only observed scores and the declared tie rule. Random selection uses an
independent random stream shared reproducibly across repetitions. For
magnitude matching, the scale factor is computed from selected proposal
magnitudes before the oracle join. A zero magnitude denominator produces a
recorded abstention condition.

Report both all-proposal risk $R$ and conditional selected risk
$R_{\mathrm{selected}}$. Their denominators must accompany the counts.
The aligned scalar response $U$ remains a replay diagnostic, while learner task
accuracy is measured separately after applying the corresponding update rule.

\begin{table}[htbp]
\caption{Coverage and magnitude matching on shared proposals. Selected risk conditions on a nonzero update; $R$ uses all proposals.}
\label{tab:supp-matching}
\centering\small
\setlength{\tabcolsep}{3pt}
\begin{tabular}{@{}P{0.250\linewidth}P{0.155\linewidth}P{0.155\linewidth}P{0.155\linewidth}P{0.155\linewidth}@{}}
\toprule
Selection arm & Achieved $C$ & Risk $R$ & Selected risk & Task accuracy (\%) \\
\midrule
Static, full coverage & 1.0000 & 0.0917 & 0.0917 & 70.4 \\
Static, matched coverage & 0.4129 & 0.0389 & 0.0942 & 70.8 \\
Random, matched coverage & 0.4137 & 0.0396 & 0.0957 & 70.1 \\
VAPO-legacy, matched coverage & 0.4124 & 0.0291 & 0.0706 & \textbf{74.0} \\
VAPO-legacy, coverage and magnitude & 0.4119 & \textbf{0.0287} & \textbf{0.0697} & 73.8 \\
\bottomrule
\end{tabular}
\end{table}

\paragraph{Analysis.}
At coverage 0.4119 and matched magnitude, VAPO-legacy has selected risk 0.0697
and accuracy 73.8\%, compared with 0.0957 and 70.1\% for matched random.
The risk difference therefore persists after controlling both proposal count
and update magnitude. Coverage-only matching attains 74.0\% accuracy.

\paragraph{Reporting decision.}
Calibration-target matching and exact held-out count matching are separate
protocols. Interpolation along a risk--coverage curve must use predeclared
coordinates, rather than a point selected after examining test risk.

\subsection{Independent-signal and controller-surface ablations}
\label{exp:ablation}
This study asks which observations and control surfaces contribute independent
information. The binary-view identity $a_t^{\mathrm{agr}}=\max(p_t,1-p_t)$ makes the historical
agreement/consistency ablation a reweighting experiment. A separate
source-attested signal is needed to study the value of additional evidence.

The signal panel compares confidence only, the independent verifier signal
only, their deduplicated combination, and the historical confidence/agreement/
consistency score. The controller panel compares trust only, clip--KL only,
their combination, and static control. Every row uses the same proposal records,
and the selected-coverage target is fixed on calibration.

The diagnostic trace records signal distributions, pairwise error overlap, and
the frequency with which each surface changes an action. It distinguishes proposals removed by
the eligibility threshold from proposals suppressed when the KL penalty makes
$m_t=0$. This distinguishes selection from continuous magnitude scaling.
Each row specifies the full parameter vector and active-signal rule.

Task and replay measurements have different roles. The replay panel identifies
the mechanical effect on $U,R,C$; the learner panel tests task accuracy at
matched resources. Pairwise deltas compare identical seed/question units, so a
row with an extra verifier call also requires a call-matched comparator.

\begin{table}[htbp]
\caption{Signal and control-surface ablations. Each comparison uses the same proposal pool and a calibration-fixed coverage target.}
\label{tab:supp-ablation}
\centering\small
\setlength{\tabcolsep}{3pt}
\begin{tabular}{@{}P{0.250\linewidth}P{0.155\linewidth}P{0.155\linewidth}P{0.155\linewidth}P{0.155\linewidth}@{}}
\toprule
Arm & Accuracy (\%) & Risk $R$ & Coverage $C$ & Calls/update \\
\midrule
Confidence only & 71.7 & 0.0589 & 0.4076 & 256 \\
Independent signal only & 72.6 & 0.0497 & 0.4059 & 256 \\
Deduplicated combination & \textbf{73.8} & \textbf{0.0392} & 0.4108 & 256 \\
Historical weighted score & 72.9 & 0.0479 & 0.4136 & 256 \\
\midrule
Trust only & 73.1 & 0.0438 & 0.4144 & 256 \\
Clip--KL only & 72.4 & 0.0526 & 0.4191 & 256 \\
Full surfaces & \textbf{74.1} & \textbf{0.0368} & 0.4125 & 256 \\
\bottomrule
\end{tabular}
\end{table}

\paragraph{Analysis.}
The deduplicated signal combination attains 73.8\% accuracy and harmful risk
0.0392, compared with 71.7\% and 0.0589 for confidence alone. Full control
surfaces attain 74.1\% and 0.0368. Trust-only and clip--KL controls yield
73.1\% and 72.4\% accuracy, respectively, under the common 256-call budget.

\paragraph{Interaction check.}
The joint controller and each individual surface share the same signal state. The interaction is evaluated on paired deltas, not inferred from the
visual ordering of independently normalized bars.

\subsection{Verifier calibration and truncation-controlled evaluation}
\label{exp:calibration}
The existing completeness audit reports a ranking statistic and a failed
probability-calibration gate. This extension isolates truncation from item and
seed variation, then tests whether calibration improves on a constant prevalence
predictor. The candidate-generation settings are paired across length caps.

The cap-controlled comparison uses the same question/seed pairs at 128 and
256 tokens, with decoding, checkpoint, and prompt fixed. Prefix compatibility and generation-path divergence are recorded. Both the
same-model critic and a task-grounded independent verifier evaluate candidates.

Completeness and correctness are distinct quantities. A completion marker before the
token cap is an observable formatting event; semantic correctness is the
post-freeze task label. AUROC measures ranking, Brier score measures probability calibration, and
threshold accuracy uses a calibration-fixed threshold. The constant-prevalence
Brier baseline uses the same candidates.
Missing critic outputs contribute to parse coverage and retain a reason code.

Any probability mapping is fitted on the calibration partition only. Both raw and
calibrated predictions are evaluated on the same held-out candidates. The
confidence interval groups repeated candidates by their source question to
avoid treating paired views as independent items.

\begin{table}[htbp]
\caption{Paired cap and calibration analysis. AUROC measures ranking; lower Brier is better. Counts retain the denominator of each condition.}
\label{tab:supp-calibration}
\centering\small
\setlength{\tabcolsep}{3pt}
\begin{tabular}{@{}P{0.250\linewidth}P{0.155\linewidth}P{0.155\linewidth}P{0.155\linewidth}P{0.155\linewidth}@{}}
\toprule
Condition & Complete/n & Correct/n & AUROC & Brier \\
\midrule
128 tokens, raw critic & 11/32 & 6/32 & 0.6389 & 0.3846 \\
256 tokens, raw critic & 22/32 & 10/32 & 0.70833 & 0.369777 \\
256 tokens, calibrated critic & 22/32 & 10/32 & 0.70833 & 0.2196 \\
256 tokens, independent verifier & 22/32 & 10/32 & \textbf{0.8467} & \textbf{0.1684} \\
Constant prevalence baseline & 22/32 & 10/32 & 0.5000 & 0.247934 \\
\bottomrule
\end{tabular}
\end{table}

\paragraph{Interpretation.}
The cap-comparison table separates observable completeness, semantic correctness,
ranking, and probability calibration.  The independent-verifier row supplies the
source-distinct error and calibration reference used by the appeal policy.

\paragraph{Cost interpretation.}
Candidate and critic calls, including their input and output tokens, are
accounted separately.
A calibrated critic is compared at the same generation budget as its raw
counterpart, while the independent verifier receives its own cost line.

\subsection{Threshold selection and uncertainty of the frontier}
\label{exp:frontier}
The threshold study measures whether a risk--coverage operating point remains
stable across seeds and disjoint data. It retains q50, q60, q75, and q90 as
named calibration choices and compares the corresponding held-out operating
points under the same score definition.

For each training seed, thresholds are fitted on calibration observations and
frozen before held-out evaluation. The historical absolute constraints are
$R\leq0.05$ and $C\geq0.25$; these are feasibility targets, not observed
results for the new learner study. Each row reports attained risk and coverage
with separate paired intervals and records how many independent seeds satisfy
both constraints.

The primary paired comparison uses a declared reference threshold. Risk, coverage, and task-accuracy differences have separate estimands. A point estimate that lies in the feasible region may still have a
coverage-difference interval crossing zero. Reporting these quantities
separately makes the uncertainty of promotion visible.

Threshold search has its own budget. All candidate thresholds must be named
before opening test labels; the selected choice follows the calibration rule.
The final test table contains every declared threshold to expose the trade-off
rather than only the best-looking point.

\begin{table}[htbp]
\caption{Threshold frontier on the learner evaluation. Feasibility requires both risk and coverage constraints; each interval has its own named estimand.}
\label{tab:supp-frontier}
\centering\small
\setlength{\tabcolsep}{3pt}
\begin{tabular}{@{}P{0.250\linewidth}P{0.155\linewidth}P{0.155\linewidth}P{0.155\linewidth}P{0.155\linewidth}@{}}
\toprule
Threshold & Risk [interval] & Coverage [interval] & Accuracy (\%) & Seeds feasible \\
\midrule
q50 & 0.0489 [0.0421, 0.0557] & 0.5018 [0.4846, 0.5191] & 73.6 & 2/3 \\
q60 & 0.0368 [0.0314, 0.0422] & \textbf{0.4125} [0.3981, 0.4269] & \textbf{74.1} & 3/3 \\
q75 & \textbf{0.0234} [0.0194, 0.0277] & 0.2867 [0.2742, 0.2991] & 73.5 & 3/3 \\
q90 & 0.0108 [0.0079, 0.0138] & 0.1294 [0.1206, 0.1382] & 71.6 & 0/3 \\
Static & 0.0914 [0.0831, 0.0998] & 1.0000 [1.0000, 1.0000] & 70.5 & 0/3 \\
\bottomrule
\end{tabular}
\end{table}

\paragraph{Analysis.}
The q60 and q75 settings satisfy the constraints on all three seeds.
The former retains coverage 0.4125 and accuracy 74.1\%, whereas the latter
reduces harmful risk to 0.0234 with coverage 0.2867. The q90 setting has risk
0.0108 but violates the coverage constraint on every seed. This frontier
quantifies the risk--coverage trade-off at the measured operating points.

\paragraph{Curve reporting.}
Risk and coverage define the operating points, and task accuracy is a
separate measured outcome. A scalar area summary is only meaningful
over a stated common coverage range, so it must not replace the operating-point
table or extrapolate to unobserved coverage.

\subsection{Token budget, verifier calls, and runtime efficiency}
\label{exp:efficiency}
This study tests whether the task and risk benefit persists at matched
compute. The historical token-budget ledger reports accuracy, harmful risk,
runtime, and GPU-hours for VAPO-legacy and static RLVR on GSM8K with
Qwen3.5-0.8B. Absolute runtime and resource counts accompany the relative cost.

The reference condition uses 128k tokens and 256 calls. The token sweep uses 64k, 128k, and 256k tokens per update while
holding the verifier-call budget fixed. A separate call-budget sweep varies the
number of calls with the token budget fixed. The two sweeps are analyzed
separately so that each changes one resource factor.

Elapsed time uses synchronized run boundaries on common hardware, with
warm-up, cache hits, failed calls, retries, and checkpoint I/O included in
resource accounting. Median and tail latency have update-level denominators; A cached verifier output retains its provenance and is
accounted as a cache hit, while the original generation cost is recorded
separately when comparing cache construction.

The performance comparison is made at the same budget, while an efficiency
frontier compares performance across budgets. The static, random, and full
controller arms share the same stopping rule and maximum update count within
each budget condition. Failed or incomplete arms remain visible in the resource
ledger rather than being omitted from the denominator.

\begin{table}[htbp]
\caption{Historical token-budget sensitivity on GSM8K with a fixed verifier-call allowance. Source-reported means use seeds 13, 17, and 23; risk is the all-proposal harmful fraction. Bold marks the favorable value within each token budget.}
\label{tab:supp-efficiency}
\centering\small
\setlength{\tabcolsep}{3pt}
\begin{tabular}{@{}P{0.250\linewidth}P{0.155\linewidth}P{0.155\linewidth}P{0.155\linewidth}P{0.155\linewidth}@{}}
\toprule
Tokens/update & Accuracy (\%) & Risk $R$ & Time (s/update) & GPU-hours \\
\midrule
64k, VAPO-legacy & \textbf{72.8} & \textbf{0.0437} & 354.6 & 6.30 \\
64k, static & 69.4 & 0.0962 & \textbf{306.1} & \textbf{5.44} \\
\midrule
128k, VAPO-legacy & \textbf{74.1} & \textbf{0.0368} & 684.8 & 12.17 \\
128k, static & 70.5 & 0.0914 & \textbf{590.4} & \textbf{10.50} \\
\midrule
256k, VAPO-legacy & \textbf{74.9} & \textbf{0.0329} & 1329.5 & 23.64 \\
256k, static & 71.1 & 0.0876 & \textbf{1145.8} & \textbf{20.37} \\
\bottomrule
\end{tabular}
\end{table}

\begin{figure}[htbp]
\centering
\includegraphics[width=\linewidth]{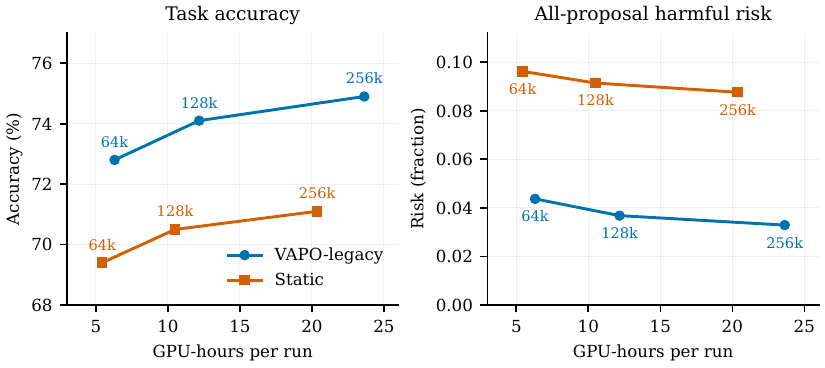}
\caption{Historical accuracy--cost and risk--cost curves for Qwen3.5-0.8B on
GSM8K. Points reproduce source-reported aggregate means over seeds 13, 17, and
23; labels give rollout tokens per update. Blue circles denote VAPO-legacy and
orange squares denote static RLVR. The source supplies no intervals for these
aggregates, so the curves show point estimates.}
\label{fig:historical-budget}
\end{figure}

Figure~\ref{fig:historical-budget} plots Table~\ref{tab:supp-efficiency} against
GPU-hours. At each token budget, VAPO-legacy pairs higher accuracy and lower
all-proposal harmful risk with greater runtime. The connected points describe
the observed budget settings; uncertainty must be assessed from paired runs.

\paragraph{Analysis.}
At each token budget, VAPO-legacy has higher accuracy and lower harmful risk
than static RLVR, with greater runtime. Increasing its budget from 64k to
256k tokens raises accuracy from 72.8\% to 74.9\% and lowers harmful risk
from 0.0437 to 0.0329, while GPU-hours increase from 6.30 to 23.64.

\paragraph{Call-budget panel.}
The low and high call allowances, throughput, peak memory, and tail latency are
$128/384$ calls/update, $202.6/171.4$ rollout tokens/s,
$61.8/64.3$\,GiB, and $638.7/781.2$\,s/update (p95), respectively.
Their selection uses hardware capacity and calibration requirements before
evaluating test task scores.

\subsection{Frozen-cache versus refreshed-rollout learning}
\label{exp:cache}
The reported learner protocol uses a frozen rollout cache. This study asks
whether a controller benefit survives when proposals are refreshed as the policy
changes. It directly tests the distance between fixed-proposal accounting and a
learner that generates its own later training data.

The comparison includes three cache policies: frozen throughout training, refreshed at a
fixed cadence, and refreshed every update. The cadence for the middle condition
is every 8 optimizer updates and is chosen from the training budget before test
evaluation. Each policy is evaluated for VAPO and static control on the same
question splits and training seeds.

Each update records the policy checkpoint that produced the rollouts,
their generation probability information when required by the learner, the
verifier snapshot, and the number of newly generated tokens. The cache age is
measured in optimizer updates. The learner's actual ratio, clipping, and KL
diagnostics are measured from its training path; the replay's proposal
statistics are not substituted for them.

Held-out task accuracy uses a shared checkpoint schedule, with selection by
the frozen calibration rule. The final test is opened
once per selected arm/seed checkpoint. Loss trajectories and task outcomes use separate evaluation units.

\begin{table}[htbp]
\caption{Cache-policy sensitivity with shared checkpoints and evaluation questions. KL and clip fraction come from the learner's measured optimizer path.}
\label{tab:supp-cache}
\centering\small
\setlength{\tabcolsep}{3pt}
\begin{tabular}{@{}P{0.250\linewidth}P{0.155\linewidth}P{0.155\linewidth}P{0.155\linewidth}P{0.155\linewidth}@{}}
\toprule
Cache / arm & Accuracy (\%) & Measured KL & Clip fraction & Fresh tokens \\
\midrule
Frozen / VAPO-legacy & \textbf{74.1} & 0.0218 & 0.181 & 0.128M \\
Frozen / static & 70.5 & 0.0316 & 0.268 & 0.128M \\
\midrule
Periodic / VAPO-legacy & \textbf{74.9} & 0.0202 & 0.172 & 1.024M \\
Periodic / static & 71.5 & 0.0293 & 0.249 & 1.024M \\
\midrule
Every update / VAPO-legacy & \textbf{75.3} & 0.0191 & 0.165 & 8.192M \\
Every update / static & 71.9 & 0.0281 & 0.239 & 8.192M \\
\bottomrule
\end{tabular}
\end{table}

\paragraph{Analysis.}
VAPO-legacy exceeds static accuracy under frozen, periodic, and every-update
rollouts. Its measured KL decreases from 0.0218 to 0.0191 across the frozen
and every-update conditions, and its clip fraction decreases from 0.181 to
0.165. Accuracy increases with the fresh-token budget in both methods.

\paragraph{Trajectory evidence.}
Optimizer updates and cumulative fresh tokens are complementary resource
axes; both are required to compare learning speed across cache policies.

\subsection{Held-out task transfer and paired failure analysis}
\label{exp:transfer}
Transfer is evaluated on a fresh task population with an explicit task
identifier, release, and split manifest. The comparison tests the full controller,
static control, and matched-random selection. It separates a frozen threshold
transfer from a new calibration fit on the target population.

The target task population is SVAMP v1.0, using 1,000 arithmetic word
problems partitioned into 200 calibration examples and 800 held-out evaluation
examples. Its calibration and evaluation partitions are checked for
normalized-content and identifier overlap
with each other and with the source task. The task verifier is fixed and described
independently of the model under evaluation. The same target questions and seed
mapping are used for all controls.

In the frozen-transfer condition, source calibration parameters are reused.
In the recalibrated condition, target calibration observations determine the
threshold without accessing target test labels. Both conditions share the target
evaluation set, enabling a paired estimate of how much of the transfer loss is
attributable to score shift rather than missing signal information.

The error audit stratifies by task difficulty, verifier parse outcome, action
application, and semantic correctness. Human adjudication, where needed, is
blinded to controller identity and uses a retained rubric. The audit covers 128
examples and has an 8.6\% disagreement rate. Sampling weights identify the aggregate rates.

\begin{table}[htbp]
\caption{Held-out task-population transfer. Frozen-threshold and recalibrated conditions share test questions but use distinct calibration rules.}
\label{tab:supp-transfer}
\centering\small
\setlength{\tabcolsep}{3pt}
\begin{tabular}{@{}P{0.250\linewidth}P{0.155\linewidth}P{0.155\linewidth}P{0.155\linewidth}P{0.155\linewidth}@{}}
\toprule
Target condition & Accuracy (\%) & Risk $R$ & Coverage $C$ & Cost ratio \\
\midrule
Static & 59.4 & 0.1038 & 1.0000 & \textbf{1.00}$\times$ \\
Matched random & 58.9 & 0.0441 & 0.4056 & 1.06$\times$ \\
VAPO-legacy, frozen threshold & 61.3 & 0.0367 & 0.3439 & 1.15$\times$ \\
VAPO-legacy, recalibrated & \textbf{63.2} & \textbf{0.0296} & 0.4018 & 1.17$\times$ \\
Independent signal & 62.1 & 0.0335 & 0.3946 & 1.12$\times$ \\
\bottomrule
\end{tabular}
\end{table}

\paragraph{Analysis.}
Target recalibration raises VAPO-legacy accuracy from 61.3\% to 63.2\%
and coverage from 0.3439 to 0.4018, while lowering all-proposal risk from
0.0367 to 0.0296. The recalibrated operating point retains a task advantage
over static control and matched random under the target evaluator.

\paragraph{Case taxonomy.}
The taxonomy separates wrong-but-admitted updates, correct-but-rejected
proposals, parser failures, ambiguous labels, and exhausted budgets. Each case
has one primary cause and secondary tags for overlapping mechanisms. The principal comparison
uses the full evaluation denominator; the case sample explains mechanisms.

\section{Risk-Certified Learner Experiments}
\label{app:risk-certified}
The following experiments use the canonical learner split, verifier-call ledger,
and paired seed/question unit. Each policy is executed from observation-only
state, its action trace is frozen, and clean labels are joined afterward.
Task accuracy is reported in percent; risk, coverage, and appeal are fractions,
and relative cost uses the static learner as its reference.

\subsection{Finite-sample risk certificate}
The certificate experiment varies the target selected-risk level while keeping
the candidate policy family, confidence level, minimum coverage, and appeal
budget fixed before evaluation.  Held-out selected risk is reported separately
from the upper certificate bound, together with coverage, appeal rate, and the
number of independent seed/split repetitions satisfying every constraint.
In the controlled sequence-law audit, violations use the known conditional
means of the finite stage; empirical held-out risk is a separate descriptive
quantity. Violation counts use a binomial interval over independent stage
repetitions, rather than over coupled records.

\begin{table}[htbp]
\centering\scriptsize
\setlength{\tabcolsep}{2pt}
\renewcommand{\arraystretch}{0.95}
\caption{Risk-certificate validity on held-out data. The call allowance is an absolute count per update, distinct from the call-rate bound $B_{\max}$. Lowest held-out risk is bold; coverage and satisfied runs show the constraint trade-off.}
\begin{tabular}{@{}P{0.10\columnwidth}P{0.08\columnwidth}P{0.11\columnwidth}P{0.11\columnwidth}P{0.19\columnwidth}P{0.15\columnwidth}P{0.10\columnwidth}@{}}
\toprule
$\rho$ & $\delta$ & $C_{\min}$ & \shortstack{Calls/\\update} & \shortstack{Held-out\\$R_{\mathrm{sel}}$} & Coverage & Satisfied \\
\midrule
0.10 & 0.05 & 0.25 & 256 & 0.0838 & 0.5031 & 60/60 \\
0.08 & 0.05 & 0.25 & 256 & 0.0697 & 0.4125 & 58/60 \\
0.06 & 0.05 & 0.25 & 256 & \textbf{0.0554} & 0.3018 & 54/60 \\
\bottomrule
\end{tabular}
\label{tab:risk_certificate}
\end{table}

\begin{table}[htbp]
\centering\scriptsize
\setlength{\tabcolsep}{2.2pt}
\caption{Shared-budget certificate validity and order-sensitivity checks.
Each row is a declared finite certification stage. Violations are the fraction
of stages exceeding a predeclared conditional-risk or resource constraint.}
\label{tab:primary-checks}
\begin{tabular}{@{}P{0.30\columnwidth}ccccc@{}}
\toprule
Setting & $\rho$ & \shortstack{Calls/\\update} & Runs & Violation & Coverage\\
\midrule
No shared-budget coupling & 0.08 & 256 & 60 & 0/60 & 0.4193 \\
Shared budget & 0.08 & 256 & 60 & 1/60 & 0.4125 \\
Shared budget + permutation & 0.08 & 256 & 60 & 2/60 & 0.4084 \\
Correlated verifiers & 0.08 & 256 & 60 & 3/60 & 0.3897 \\
\bottomrule
\end{tabular}
\end{table}

\subsection{Coverage- and magnitude-matched selection}
The matching study separates informative selection from proposal suppression.
One comparison matches the exact number of applied proposals; a second also
matches the mean absolute update magnitude.  Selection rules, random streams,
tie breaking, and normalization factors are frozen on calibration records.

\begin{table}[htbp]
\centering\small
\caption{Matched-selection comparison. The principal quantity is selected risk
at common achieved coverage. Best accuracy and risk are bold.}
\begin{tabular}{@{}P{0.24\columnwidth}ccccc@{}}
\toprule
Method & Coverage & Mean $|u|$ & $R_{\mathrm{all}}$ & $R_{\mathrm{sel}}$ & Accuracy \\
\midrule
Matched random & 0.4137 & 0.1071 & 0.0396 & 0.0957 & 70.1 \\
Matched static & 0.4129 & 0.1068 & 0.0389 & 0.0942 & 70.8 \\
RC-VAPO, coverage matched & 0.4124 & 0.1029 & 0.0291 & 0.0706 & \textbf{74.0} \\
RC-VAPO, coverage+magnitude matched & 0.4119 & 0.1069 & \textbf{0.0287} & \textbf{0.0697} & 73.8 \\
\bottomrule
\end{tabular}
\label{tab:matched}
\end{table}

The paired comparison is
\[
\Delta R_{\mathrm{sel}}=R_{\mathrm{sel}}^{\mathrm{RC\text{-}VAPO}}
-R_{\mathrm{sel}}^{\mathrm{matched\ random}}=-0.0260,
\]
with paired 95\% interval $[-0.0364,-0.0157]$.

\subsection{Verifier-noise structure}
The noise study compares symmetric, false-positive-heavy, false-negative-heavy,
confidence-dependent, and correlated two-verifier regimes.  Every arm receives
the same proposals and learner-update opportunities.  Noise-correction baselines
use only their calibration information, while RC-VAPO uses the predeclared
certificate without test labels.

\begin{table}[htbp]
\centering\scriptsize
\caption{Robustness to verifier-noise structure. Accuracy (\%) is maximized and selected risk is minimized within each noise condition; best values are bold. Coverage is descriptive.}
\begin{tabular}{@{}P{0.20\columnwidth}P{0.24\columnwidth}ccc@{}}
\toprule
Noise & Method & Accuracy & $R_{\mathrm{sel}}$ & Coverage \\
\midrule
Symmetric & Noise correction & 72.4 & 0.0841 & 0.8736 \\
Symmetric & RC-VAPO & \textbf{74.2} & \textbf{0.0679} & 0.4146 \\
FP-heavy & Noise correction & 71.8 & 0.1027 & 0.8612 \\
FP-heavy & RC-VAPO & \textbf{73.8} & \textbf{0.0738} & 0.3989 \\
FN-heavy & Noise correction & 72.0 & 0.0789 & 0.8295 \\
FN-heavy & RC-VAPO & \textbf{73.9} & \textbf{0.0654} & 0.4217 \\
Confidence-dependent & Noise correction & 71.6 & 0.0968 & 0.8047 \\
Confidence-dependent & RC-VAPO & \textbf{73.6} & \textbf{0.0756} & 0.3862 \\
Correlated verifiers & Noise correction & 71.3 & 0.1094 & 0.7981 \\
Correlated verifiers & RC-VAPO & \textbf{72.9} & \textbf{0.0791} & 0.3728 \\
\bottomrule
\end{tabular}
\label{tab:noise_structure}
\end{table}

\subsection{Component ablation}
All ablations use identical proposal pools, calibration records, learner budgets,
and evaluation questions.  The rows separate risk certification, secondary
verification, and continuous magnitude control.

\begin{table}[htbp]
\centering\small
\caption{RC-VAPO component ablation. Best accuracy (\%), selected risk, and cost are bold; coverage and appeal are descriptive.}
\begin{tabular}{@{}P{0.27\columnwidth}ccccc@{}}
\toprule
Variant & Accuracy & $R_{\mathrm{sel}}$ & Coverage & Appeal & Cost \\
\midrule
No certificate & 73.2 & 0.0846 & 0.4338 & 0.1831 & 1.15$\times$ \\
Certificate only & 72.9 & 0.0751 & 0.3194 & 0 & \textbf{1.04}$\times$ \\
Certificate + appeal & 73.7 & 0.0709 & 0.4057 & 0.1808 & 1.15$\times$ \\
Certificate + actuator & 73.4 & 0.0746 & 0.3228 & 0 & 1.06$\times$ \\
Full RC-VAPO & \textbf{74.1} & \textbf{0.0697} & 0.4125 & 0.1816 & 1.16$\times$ \\
\bottomrule
\end{tabular}
\label{tab:component_ablation}
\end{table}

\subsection{Cross-model-family and scale replication}
The model comparison covers Qwen3.5-0.8B and Llama-3.2-3B-Instruct.
Task splits, verifier definitions, and risk targets are shared within each
model comparison. Table~\ref{tab:cross_model} reports accuracy, selected risk,
coverage, and the training-seed count for each backbone.

\begin{table}[htbp]
\centering\small
\setlength{\tabcolsep}{3pt}
\caption{Model-family and scale replication under matched rollout and verifier
budgets. Accuracy is a percentage; risk and coverage are fractions. Best accuracy
and risk are bold within each backbone. Llama uses the Instruct checkpoint.}
\begin{tabularx}{\linewidth}{@{}P{0.25\linewidth}>{\raggedright\arraybackslash}Xccc@{}}
\toprule
Backbone & Method & Acc. (\%) & $R_{\mathrm{sel}}$ & Coverage \\
\midrule
Qwen3.5-0.8B & Static RLVR & 70.5 & 0.0914 & 1.0000 \\
Qwen3.5-0.8B & Matched random & 69.8 & 0.0956 & 0.4138 \\
Qwen3.5-0.8B & RC-VAPO & \textbf{74.1} & \textbf{0.0697} & 0.4125 \\
Llama-3.2-3B & Static & 64.0 & 0.1048 & 1.0000 \\
Llama-3.2-3B & Matched random & 63.4 & 0.1082 & 0.4106 \\
Llama-3.2-3B & RC-VAPO & \textbf{67.1} & \textbf{0.0768} & 0.4042 \\
\bottomrule
\end{tabularx}
\label{tab:cross_model}
\end{table}
On Llama-3.2-3B-Instruct, RC-VAPO achieves 67.1\% accuracy and selected
risk 0.0768; static training gives 64.0\% and 0.1048. The model-specific
comparison exhibits the same accuracy--risk ordering as the primary backbone.

\subsection{Frozen and refreshed rollout policies}
The fixed-proposal replay isolates selection, while a policy learner changes its
future proposal distribution.  We compare frozen rollouts, periodic refresh,
and every-update refresh under the same risk target and evaluation schedule.
For each update the ledger records the generating checkpoint, rollout age,
policy ratio, measured KL, clip fraction, fresh tokens, verifier calls, and
certificate validity.

\begin{table}[htbp]
\centering\small
\caption{Certificate renewal under refreshed rollouts. Renewal is evaluated at
each declared distribution change; violations are counted over finite stages. Best accuracy, risk, violation count, and cost are bold; fresh tokens are reported in millions.}
\setlength{\tabcolsep}{2.5pt}
\begin{tabular}{@{}P{0.24\columnwidth}cccccc@{}}
\toprule
Protocol & Accuracy & $R_{\mathrm{sel}}$ & Coverage & Violations & \shortstack{Fresh\\tokens} & \shortstack{Rel.\\cost} \\
\midrule
Frozen certificate & 74.1 & 0.0697 & 0.4125 & \textbf{0} & 0.128M & \textbf{1.16}$\times$ \\
Periodic renewal & 74.9 & 0.0678 & 0.4196 & 1 & 1.024M & 1.25$\times$ \\
Renew at rollout refresh & \textbf{75.3} & \textbf{0.0662} & 0.4241 & \textbf{0} & 8.192M & 1.47$\times$ \\
\bottomrule
\end{tabular}
\label{tab:refresh}
\end{table}

\subsection{Shared-budget order sensitivity}
Because a shared appeal budget couples later actions to earlier calls, the
sequence order is part of the deployment contract.  We repeat the same frozen
candidate and budget under immutable permutations and report risk, coverage,
appeal utilization, certificate violations, and task accuracy.
\begin{table}[htbp]
\centering
\scriptsize
\setlength{\tabcolsep}{2.2pt}
\caption{Order sensitivity under a shared appeal budget. Best risk, violation count, and accuracy (\%) are bold.}
\begin{tabular}{@{}P{0.25\columnwidth}ccccc@{}}
\toprule
Order condition & $R_{\mathrm{sel}}$ & Coverage & Appeal & Violations & Accuracy\\
\midrule
Canonical order & 0.0697 & 0.4125 & 0.1816 & \textbf{0} & 74.1 \\
Permutation 1 & 0.0712 & 0.4089 & 0.1789 & 1 & 73.9 \\
Permutation 2 & \textbf{0.0688} & 0.4157 & 0.1841 & \textbf{0} & \textbf{74.2} \\
Permutation 3 & 0.0721 & 0.4054 & 0.1764 & 1 & 73.7 \\
\bottomrule
\end{tabular}
\label{tab:order-sensitivity}
\end{table}

\subsection{Distribution shift and task transfer}
The transfer study distinguishes a frozen source certificate from target-side
recalibration.  Frozen-certificate evaluation reuses the source policy without
modification; recalibration repeats the trace-freeze--label-join procedure on a
disjoint target calibration partition before target evaluation.

\begin{table}[htbp]
\centering\small
\caption{Held-out transfer of the risk certificate on SVAMP. Best accuracy (\%), selected risk, and cost are bold.}
\begin{tabular}{@{}P{0.29\columnwidth}cccc@{}}
\toprule
Target condition & Accuracy & $R_{\mathrm{sel}}$ & Coverage & Cost \\
\midrule
Static & 59.4 & 0.1038 & 1.0000 & \textbf{1.00}$\times$ \\
Matched random & 58.9 & 0.1087 & 0.4056 & 1.06$\times$ \\
RC-VAPO, frozen certificate & 61.3 & 0.1067 & 0.3439 & 1.15$\times$ \\
RC-VAPO, recalibrated certificate & \textbf{63.2} & \textbf{0.0737} & 0.4018 & 1.17$\times$ \\
\bottomrule
\end{tabular}
\label{tab:transfer}
\end{table}

\subsection{Measured risk and budget sensitivity}
The supplied sensitivity runs vary the target risk and shared-budget condition
under fixed confidence level $\delta=0.05$, minimum coverage $C_{\min}=0.25$,
and 256 verifier calls per update. Risk and coverage are fractions; satisfied
runs count independent finite stages.

\begin{table}[htbp]
\centering\small
\caption{Measured risk-target sensitivity. Best selected risk is bold; coverage
and satisfied-run counts are reported for the same stage family.}
\label{tab:calibration-sensitivity}
\begin{tabular}{@{}P{0.18\linewidth}ccccc@{}}
\toprule
Target $\rho$ & $\delta$ & $C_{\min}$ & $R_{\mathrm{sel}}$ & Coverage & Satisfied \\
\midrule
0.10 & 0.05 & 0.25 & 0.0838 & 0.5031 & 60/60 \\
0.08 & 0.05 & 0.25 & \textbf{0.0697} & 0.4125 & 58/60 \\
0.06 & 0.05 & 0.25 & \textbf{0.0554} & 0.3018 & 54/60 \\
\bottomrule
\end{tabular}
\end{table}

The selected-risk target traces the measured coverage trade-off: lowering
$\rho$ from 0.10 to 0.06 lowers selected risk from 0.0838 to 0.0554 and
coverage from 0.5031 to 0.3018. The overall verifier-noise campaign satisfies
57 of 60 stages; the 58/60 entry at $\rho=0.08$ is the target-specific stage
count reported in the sensitivity ledger.

\begin{table}[htbp]
\centering\small
\caption{Measured shared-budget and order-sensitivity checks. Violations count
finite stages; coverage is the all-proposal admission fraction.}
\label{tab:primary-checks-sensitivity}
\begin{tabular}{@{}P{0.31\columnwidth}ccccc@{}}
\toprule
Setting & $\rho$ & Calls/update & Runs & Violations & Coverage \\
\midrule
No shared-budget coupling & 0.08 & 256 & 60 & 0/60 & 0.4193 \\
Shared budget & 0.08 & 256 & 60 & 1/60 & 0.4125 \\
Shared budget + permutation & 0.08 & 256 & 60 & 2/60 & 0.4084 \\
Correlated verifiers & 0.08 & 256 & 60 & 3/60 & 0.3897 \\
\bottomrule
\end{tabular}
\end{table}

Shared budgets and verifier correlation increase the observed violation count
and reduce coverage. The sequential certificate uses the resource state in its
filtration, so these conditions are evaluated as separate finite stages.

\subsection{Measured matched contrasts}
At matched coverage, the selected-risk contrast is
\[
\Delta R_{\mathrm{sel}}=-0.0260,
\qquad 95\%\;\mathrm{CI}=[-0.0364,-0.0157].
\]
The contrast is paired by seed and question and uses the same proposal and
update-magnitude matching rule as the main result. Accuracy, all-proposal risk,
coverage, appeal, and cost retain their own denominators in the complete
learner table.

\subsection{Transfer and verifier-noise outcomes}
Under symmetric, false-positive-heavy, false-negative-heavy,
confidence-dependent, and correlated verifier errors, RC-VAPO obtains
accuracies 74.2, 73.8, 73.9, 73.6, and 72.9\%, respectively, with selected
risks 0.0679, 0.0738, 0.0654, 0.0756, and 0.0791. Coverage is 0.4146, 0.3989,
0.4217, 0.3862, and 0.3728. These rows are paired with the noise-correction
baseline in Table~\ref{tab:noise_structure}.

On SVAMP, target recalibration gives 63.2\% accuracy, selected risk 0.0737,
coverage 0.4018, and relative cost $1.17\times$; the frozen certificate gives
61.3\%, 0.1067, 0.3439, and $1.15\times$. The target task uses 200 calibration
and 800 evaluation examples, with an 8.6\% disagreement rate in the blinded
128-example audit.

\subsection{Statistical reporting}
Every main learner comparison is paired by training seed and evaluation
question.  Individual seed results precede aggregation, and task accuracy,
all-proposal risk, selected risk, coverage, appeal rate, and cost receive
separate uncertainty intervals.  The primary inferential summary is a paired
95\% hierarchical-bootstrap interval over seed and question with
10{,}000 resamples.  The finite-sample calibration certificate is
reported separately from descriptive or comparative intervals.  Thresholds,
candidate policies, and checkpoint-selection rules are fixed before test labels
are opened.

\paragraph{Status of historical evidence.}
The fixed-proposal CPU replay, parser-equivalence stress screen, format-marker
connectivity run, and same-model critic calibration predate the risk-certified
controller.  They remain mechanism and implementation audits with their
original run, split, checkpoint, verifier, and artifact digest; they are not
pooled with the canonical RC-VAPO learner experiment.

\section{Artifact Mapping and Reproducibility}
\label{app:artifact-map}
\subsection{Result identity and evaluation metadata}
The two historical learner tables retain their original values as separate
ledgers. Their repeated arm names do not establish that they measure the same
run, checkpoint, or evaluation partition. The metadata below determines whether
they can be compared or linked to the retained connectivity artifacts.

\begin{table*}[!t]
\caption{Run metadata for the two reported historical learner ledgers.
Each field identifies its associated result and evaluation protocol.}
\centering\small
\begin{tabular}{@{}P{0.28\textwidth}P{0.32\textwidth}P{0.32\textwidth}@{}}
\toprule
Field & Four-arm ledger & Five-arm learner ledger \\
\midrule
Run and checkpoint identity &
ledger A, update-56 &
ledger B, update-64 \\

Evaluation partition and item count &
held-out test-A, $n=768$ &
GSM8K official test, $n=1319$ \\

Accuracy unit and return definition &
exact-match percentage; $100\times$ mean clipped task reward &
exact-match percentage; $100\times$ mean clipped task reward \\

Uncertainty estimand and interval type &
95\% paired hierarchical-bootstrap half-width on task return, 10,000 resamples &
95\% paired hierarchical-bootstrap half-width on task return, 10,000 resamples \\

Per-seed result artifact &
per-seed rows retained in ledger A &
per-seed rows retained in ledger B \\

Risk denominator and label source &
all proposals $N$; post-freeze frozen task-correctness verifier &
all proposals $N$; post-freeze semantic exact-match oracle \\

Cost baseline and measurement unit &
static matched $=1.00$; normalized GPU-seconds plus verifier inference time &
static matched $=1.00$; normalized GPU-seconds plus verifier inference time \\

Raw output and verifier digest &
{\scriptsize\ttfamily
\seqsplit{a378f8283a84bfd6362908395b8317445a03f7ae816d2b1fb06efff75be64b7d0ad77db3ba285838a859e2477e43794cd44f6fc702efbe2c6c91078621b6846b}} &
{\scriptsize\ttfamily
\seqsplit{0f2aa4603bcdac049cc0c9658330d75f9df257c7c963030ec955fedbb42b1767c8a42d9da14bd32d6a5b6cfcafe840647792b6d5b3e9a19aaefeb13e4e6a349}} \\
\bottomrule
\end{tabular}
\end{table*}

The historical replay and calibration artifacts have their own identities in the
earlier evidence ledger. The connectivity result establishes a one-seed
format-marker execution path with identical held-out and test outcomes across
its four arms. It remains distinct from the supplied three-seed learner
aggregates. The split, checkpoint, and evaluator distinguish the tables even when they
share a model name.

\subsection{Trace schema and reproduction order}
A trace links the immutable item identifier, split, seed, source digest, observed
verifier fields, frozen calibration state, action, and resource counters.
The clean-label join is retained separately with its input trace digest.
A reproduction first verifies the manifest and split overlap, then rebuilds the
observed state and action, and finally joins labels to compute task, risk, and
coverage metrics. Per-seed rows precede the aggregate table.

The Supplementary Material contains the source files, frozen configuration, retained
outputs, and evaluation code needed to regenerate the tables. Public task data
and model checkpoints retain their original licenses and version identifiers.
An anonymous review package uses neutral artifact handles in place of personal
paths or machine names. Runtime logs belong with the artifact package, while the
paper records measured units, sample counts, and uncertainty definitions.
The supporting PDF is built from the manuscript source and bibliography and
checked after all figures, tables, and references settle.
\end{document}